\documentclass[11pt]{article}

\usepackage{acl}
\usepackage{times}
\usepackage{latexsym}
\usepackage[T1]{fontenc}
\usepackage[utf8]{inputenc}
\usepackage{microtype}
\usepackage{inconsolata}
\usepackage{graphicx}
\usepackage{booktabs}
\usepackage{multirow}
\usepackage{amsmath}
\usepackage{amssymb}
\usepackage{xcolor}
\usepackage{enumitem}
\usepackage{pifont}
\usepackage{hyperref}
\usepackage{pifont}
\usepackage{fontawesome5}

\usepackage{booktabs}
\usepackage{tabularx}
\usepackage{enumitem}
\usepackage{tcolorbox}
\tcbuselibrary{skins, breakable}

\newcommand{\cmark}{\ding{51}}
\newcommand{\xmark}{\ding{55}}

\title{Evaluating and Calibrating LLM Confidence on Questions with Multiple Correct Answers}

\author{
Yuhan Wang\textsuperscript{\rm 1,2} \thanks{~~Work done during an internship at ICT,CAS} \footnotemark[2] \quad
Shiyu Ni\textsuperscript{\rm 1,2,3} \thanks{~~Equal contributions} \quad \\
\textbf{Zhikai Ding}\textsuperscript{\rm 1,2} \footnotemark[1] \quad
\textbf{Zihang Zhan}\textsuperscript{\rm 4} \footnotemark[1] \quad
\textbf{Yuanzi Li}\textsuperscript{\rm 5} \quad
\textbf{Keping Bi}\textsuperscript{\rm 1,2,3} \thanks{~~Corresponding author} \\
\textsuperscript{\rm 1} State Key Laboratory of AI Safety \\
\textsuperscript{\rm 2} Institute of Computing Technology, Chinese Academy of Sciences \\
\textsuperscript{\rm 3} University of Chinese Academy of Sciences \\
\textsuperscript{\rm 4} Tsinghua University \quad
\textsuperscript{\rm 5} Renmin University of China \\
\texttt{fzzh040114@gmail.com} \quad
\texttt{\{nishiyu23z,bikeping\}@ict.ac.cn} \\
\href{https://github.com/Trustworthy-Information-Access/Calibration-Under-Multiple-Correct-Answers}{\faGithub~Code} 
\quad
  \href{https://huggingface.co/datasets/fzzh/MACE}{\includegraphics[height=1.0em]{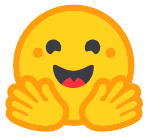}~Datasets} 
}

\begin{document}
\maketitle

\begin{abstract} 
Confidence calibration is essential for making large language models (LLMs) reliable, yet existing training-free methods have been primarily studied under single-answer question answering. 
In this paper, we show that these methods break down in the presence of multiple valid answers, where disagreement among equally correct responses leads to systematic underestimation of confidence. To enable a systematic study of this phenomenon, we introduce MACE, a benchmark of 12,000 factual questions spanning six domains with varying numbers of correct answers. Experiments across 15 representative calibration methods and four LLM families (7B–72B) reveal that while accuracy increases with answer cardinality, estimated confidence consistently decreases, causing severe miscalibration for questions with mixed answer counts. Notably, larger models are more severely affected, as they internalize substantially more knowledge. To address this issue, we propose Semantic Confidence Aggregation (SCA), which aggregates confidence over multiple high-probability sampled responses. SCA achieves state-of-the-art calibration performance under mixed-answer settings while preserving strong calibration on single-answer questions.
\end{abstract}

\begin{figure*}[t] 
    \centering
    \includegraphics[width=\linewidth]{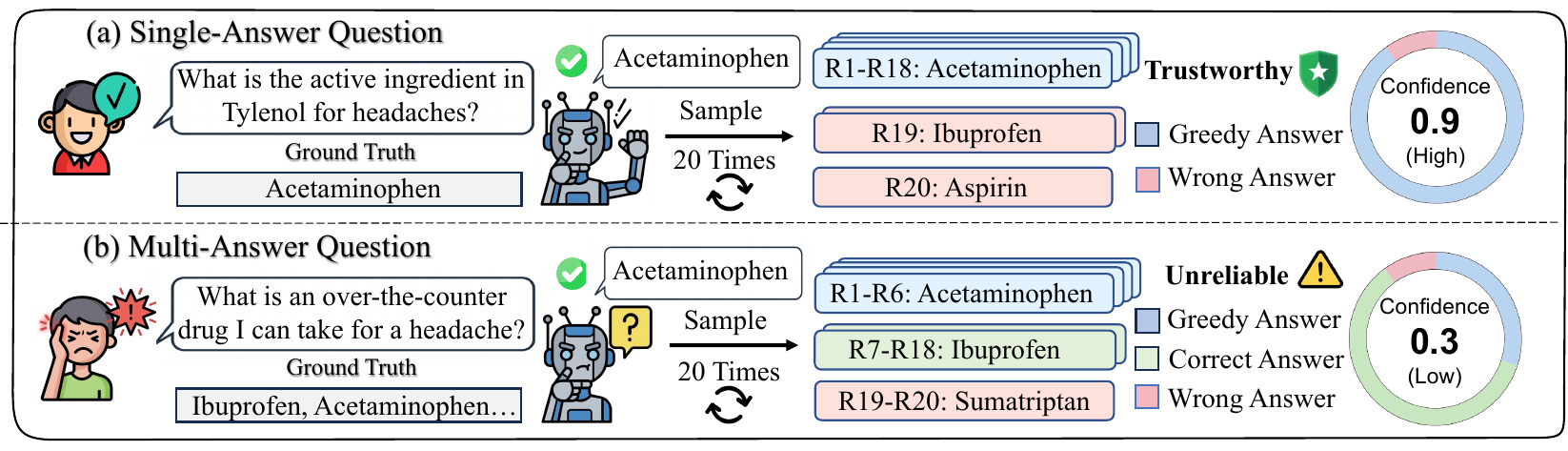}
\caption{
\textbf{Failure of consistency-based calibration on multi-answer questions. }Disagreement among sampled correct answers yields low estimated confidence, 
indistinguishable from the low confidence caused by the presence of incorrect answers, thereby making correct responses appear unreliable.
}
\label{fig:case}
\end{figure*}


\section{Introduction}
Confidence calibration aims to align a model’s expressed confidence with its true predictive performance~\citep{guo2017calibration, desai2020calibration}, and has long been a central research topic for improving the reliability of machine learning systems. In recent years, confidence calibration for large language models (LLMs) has attracted increasing attention, as honesty has been recognized as a fundamental aspect of aligning LLMs with human values~\citep{askell2021general}. When properly calibrated, LLMs’ confidence can serve as a meaningful indicator of their uncertainty, reflecting the likelihood of hallucination~\citep{Maynez2020Faithfulness, Min2023FActScore} and guiding when retrieval augmentation is required~\citep{ni2024llms}.

Confidence calibration methods can be divided into training-based and training-free approaches. Training-based methods learn to align confidence with ground-truth accuracy and can achieve strong in-domain calibration ~\citep{lin2022teaching, ni2025towards}. However, they require substantial annotation and often generalize poorly to other questions or domains ~\citep{ni2025annotation}. In this paper, we focus on training-free methods, which can be roughly categorized into four approaches:
1) leveraging token-level generation probabilities~\citep{guo2017calibration, desai2020calibration,jiang2021can,si2022prompting}, 
2) prompting models to explicitly verbalize confidence~\cite{lin2022teaching, yin2023large,xiong2023can,ni2024llms}, 
3) measuring semantic consistency across multiple sampled responses~\citep{Manakul2023SelfCheckGPTZB, kuhn2023semantic,zhang2023sac3},
and
4) employing the model itself to evaluate whether its generated answer is correct~\citep{kadavath2022language, tian2023just}.

Among these, response-consistency–based methods achieve state-of-the-art (SOTA) performance. The underlying assumption is that greater diversity among sampled responses signals higher model uncertainty—if the model "knows" the answer, its outputs should converge. This assumption is reasonable when each question admits a single correct answer, as divergent outputs genuinely reflect the model's hesitation.
However, this assumption breaks down when a question has multiple valid answers. As illustrated in Figure~\ref{fig:case}, consider two questions that both yield 18 out of 20 correct sampled responses. For the single-answer question "What is the active ingredient in Tylenol for headaches?" (panel a), the 18 correct responses all converge on the same answer—Acetaminophen—and the consistency-based confidence estimate is a trustworthy 0.9. For the multi-answer question "What is an over-the-counter drug I can take for a headache?" (panel b), however, the 18 correct responses split across equally valid answers—Acetaminophen (R1–R6) and Ibuprofen (R7–R18). Despite achieving the same accuracy, this natural variation among correct answers drives the estimated confidence down to  0.3, and the model is flagged as unreliable. The model suffers a severe trust crisis not because it is wrong, but because it is right in more than one way.

Existing benchmarks largely overlook the multi-answer setting and are not directly suitable for evaluating confidence calibration under it (see Appendix~\ref{sec:limitation-benchmarks} for a detailed discussion).
To systematically study confidence estimation under multi-answer question answering (QA), we introduce MACE (\textbf{M}ulti-\textbf{A}nswer \textbf{C}onfidence \textbf{E}stimation), a benchmark of 12,000 question–answer pairs spanning six factual domains, where each question has 1, 2, 4, or 6 correct answers. \looseness=-1

Using MACE, we evaluate 15 representative confidence calibration methods across four LLM families (LLaMA, Qwen, DeepSeek, and GPT), ranging from 7B to 72B parameters. Our experiments reveal that: (1) as the number of correct answers increases, LLMs achieve higher QA accuracy but exhibit lower confidence; (2) in realistic settings with mixed ground-truth answer cardinalities, methods that achieve SOTA calibration on single-answer questions collapse due to severe miscalibration on multi-answer questions; and (3) this confidence degradation is more pronounced for larger models (e.g., 70B), as they tend to alternate among a wider set of correct answers than smaller models. \looseness=-1

 

To address this, we propose SCA (\textbf{S}emantic \textbf{C}onfidence \textbf{A}ggregation), which calibrates confidence by aggregating the probabilities of multiple high-confidence answers rather than relying on the most confident one. Specifically, we use the token-level sequence generation probability of each sampled response as its confidence signal. Since low-confidence responses contribute minimally, a simple summation over sampled responses yields improved calibration under mixed-answer settings, outperforming SOTA baselines while preserving strong performance on single-answer questions.

In summary, this work advances confidence calibration from single-answer QA to a more general setting with mixed numbers of ground-truth answers. We introduce the MACE benchmark, conduct a comprehensive evaluation across four LLM families, and propose SCA, which achieves state-of-the-art calibration performance in this general QA scenario. \looseness=-1

\section{Related Work}
\label{sec:related}

\paragraph{Calibration in Large Language Models.}
Existing calibration methods for large language models (LLMs) can be training-based and training-free. Training-based methods require finetuning to enhance knowledge boundary perception \citep{lin2022teaching, zhang2024r, yang2024alignment}.
Training-free methods can be broadly categorized into four approaches:
\textit{Probability-based } methods leverage the model's internal representations, such as token probabilities, logits, or hidden activations \citep{jiang2021can, kadavath2022language, ren2022out, wightman2023strength, ling2024uncertainty,ni2025towards}. 
\textit{Verbalized-based} methods prompt LLMs to directly express confidence in natural language \citep{lin2022teaching, tian2023just, tanneru2024quantifying}.
\textit{Consistency-based} methods estimate reliability by measuring agreement across multiple generations or reasoning chains
\citep{Manakul2023SelfCheckGPTZB, cole2023selectively, lyu2025calibrating, ni2025annotation}.
\textit{Self-evaluation-based} methods ask the model to judge its own correctness \citep{kadavath2022language, tian2023just}.


\paragraph{Datasets for Confidence Estimation.}
Existing benchmarks can be categorized by answer-space structure.
\textit{Multiple-choice} benchmarks include MMLU-PRO~\citep{wang2024mmlu}, ARC~\citep{clark2018think}, and SATA-Bench~\citep{xu2025sata}.
\textit{Open single-answer QA} benchmarks include TriviaQA~\citep{joshi2017triviaqa}, Natural Questions~\citep{kwiatkowski2019natural}, and Mintaka~\citep{sen2022mintaka}.
\textit{Open multi-answer QA} benchmarks include WebQuestions~\citep{berant2013semantic} and ComplexWebQuestions~\citep{talmor2018web}.
However, none simultaneously provides exhaustive ground truth, multi-answer questions, and controlled answer cardinality—three properties essential for studying calibration under probability mass split (see Appendix~\ref{sec:limitation-benchmarks} for a detailed analysis). To fill this gap, we introduce the benchmark \textbf{MACE}.

\begin{figure*}[t]
  \centering
  \includegraphics[width=1\textwidth]{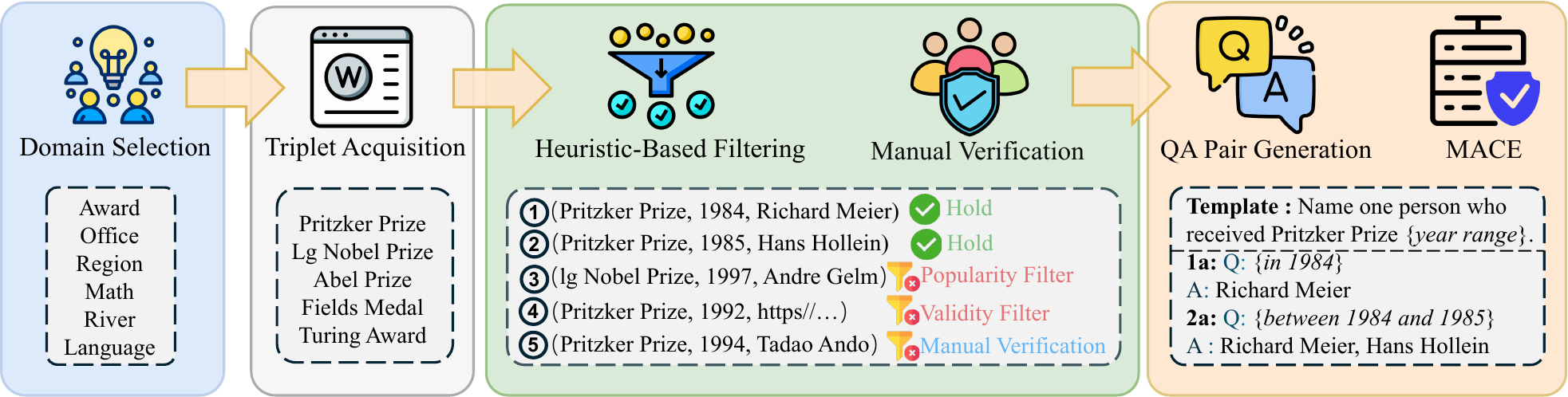}
\caption{\textbf{The MACE benchmark construction pipeline}, illustrated using the \textit{Award} domain. 
\ding{172}\ding{173} \textbf{Valid triplets} (e.g., Pritzker Prize winners) are retained.
\ding{174} \textbf{Low-popularity triplets} (e.g., obscure awards like the Ig Nobel Prize) are removed via \textit{Popularity Filter}.
\ding{175} \textbf{Noisy triplets} containing invalid formats (e.g., URLs) are removed via a \textit{Validity Filter}.
\ding{176} \textbf{Factually incorrect triplets} are identified and removed during the \textit{Manual Verification}.
Finally, QA pairs are generated using triplets under four ground-truth counts settings (1a, 2a, 4a, 6a).}
  \label{fig:data_structure}
\end{figure*}


\section{The MACE Benchmark}
\label{sec:dataset}

In this section, we provide the overview and construction process of the MACE  benchmark.
\newcommand{\domainAward}{Honorable Award}
\newcommand{\domainOffice}{Political Office}
\newcommand{\domainCountry}{Regional Affiliation}
\newcommand{\domainMath}{Mathematical Concept}
\newcommand{\domainRiver}{Natural River}
\newcommand{\domainLanguage}{Linguistic Culture}

\newcommand{\Award}{Award}
\newcommand{\Office}{Office}
\newcommand{\Country}{Region}
\newcommand{\Math}{Math}
\newcommand{\River}{River}
\newcommand{\Language}{Language}

\newcommand{\answerspace}{knowledge coverage}

\subsection{Overview}
To study how existing confidence calibration methods perform on multi-answer questions, we introduce MACE (\textbf{M}ulti-\textbf{A}nswer \textbf{C}onfidence \textbf{E}stimation), a benchmark comprising questions with one or multiple correct answers. 
MACE spans six factual domains: \domainAward~(\Award), \domainOffice~(\Office), \domainCountry~(\Country), \domainMath~(\Math), \domainRiver~(\River), and \domainLanguage~(\Language).
For each domain, MACE includes four question types with 1, 2, 4, or 6 correct answers, respectively (abbreviated as 1a, 2a, 4a, and 6a, in the paper), with 500 questions per type. 
The construction of MACE has three key phases, shown in Figure~\ref{fig:data_structure}: \textbf{knowledge collection} that collects knowledge triplets $\{\text{subject}, \text{relation}, \text{object}\}$ in the domains, \textbf{knowledge filtering} that filters out low-quality triplets, and \textbf{QA pair generation} that constructs natural language QA pairs based on the triplets. 
Dataset statistics are summarized in Table~\ref{tab:dataset}. \looseness=-1

\begin{table}[t]
\centering
\small 
\caption{
\textbf{Statistics of the six factual domains in MACE.}
\textbf{Subj.}: number of subjects collected from Wikidata;
\textbf{Obj.}: number of distinct objects associated with the subjects;
\textbf{Const-Q}: number of questions constructed from subject–object pairs;
\textbf{Used-Q}: number of used questions  via clustering-based stratified sampling.
}
\label{tab:dataset}
\begin{tabular}{lrrrr}
\toprule
\textbf{Domain} & \textbf{Subj.} & \textbf{Obj.}  & \textbf{Const-Q} & \textbf{Used-Q} \\
\midrule
Award    & 175 & 1,634 & 9,384  & 2,000 \\
Office   & 159 & 1,706 & 6,177  & 2,000 \\
Region   & 99  & 1,074 & 5,120  & 2,000 \\
Math     & 5   & 2,957 & 10,000 & 2,000 \\
River    & 781 & 114   & 10,000 & 2,000 \\
Language & 847 & 141   & 10,000 & 2,000 \\
\bottomrule
\end{tabular}
\end{table}

\subsection{Knowledge Collection}
\label{sec:data-acquisition}

\paragraph{Domain Identification.} 
The challenge in data construction lies in identifying questions that allow multiple valid answers while remaining complete. For example, for the question "Tell me a great scientist in 2025,” it is impractical to enumerate all correct answers.
To ensure that answers are clear and complete, we begin by collaborating with LLMs to identify suitable domains. Balancing coverage and annotation cost, we ultimately select six domains (See Table~\ref{tab:dataset}). Details are provided in Appendix~\ref{app:domain-identification}.


\paragraph{Triplet Collection.}
In this step, we collect knowledge for each domain.
For the Math domain, knowledge is derived via predefined knowledge types, such as identifying prime numbers.
For other domains, following prior research~\cite{sen2022mintaka,yuksekgonul2023attention,mallen2023not}, we collect knowledge triplets from Wikidata\footnote{https://query.wikidata.org/sparql} which is a widely recognized knowledge source. 
Knowledge triplets are structured in the form of (subject, relation, object). For a given domain, we first retrieve all relevant subjects, and then query the corresponding objects.
The subject type, relation, and object type of each domain can be found in Table~\ref{tab:domain-schema}. The subject counts are reported in the \texttt{Raw} column of Table~\ref{tab:filter}. More details are described in ~\S~\ref{sec:triplet_collection}.

\subsection{Knowledge Filtering}
\label{sec:Data Filtering}
To ensure answer correctness and completeness, we conduct heuristic filtering and manual verification. The count of the remaining subjects after each step is shown in Table~\ref{tab:filter}. 

We use a two-stage pipeline to remove low-quality data:
(i) \textit{Popularity Filter}: retains only high-traffic subjects based on Wikipedia pageviews;
(ii) \textit{Validity Filter}: further eliminates invalid or noisy subjects via domain-specific rules. Details can be found in Appendix~\ref{appendix:dataset-filtering}.
This step filters 92.4\% of the data, retaining a high-quality subset and substantially reducing the workload for human verification. \looseness=-1

\paragraph{Manual Verification.} We recruit ten domain experts to manually audit the remaining triplets, recognizing any remaining factual errors and enriching incomplete entries. Finally, this step removes an additional 19.05\% of subjects, achieving high inter-annotator agreement (Cohen’s $\kappa = 0.94$). Details can be found in~\S~\ref{sec:human_check}.


\subsection{QA Pair Generation}
\label{sec:qa-synthesis}
Following previous research~\cite{mallen2023not,yuksekgonul2023attention}, we construct QA pairs using the collected knowledge triplets.
For each domain, we create questions based on a subject and its associated relation, with the object serving as the answer. By varying the subjects and relations, we can generate questions with different numbers of correct answers, e.g., ``Name one person who received \{award\} between \{lower\_year\} and \{upper\_year\}''. Details can be found in~\S~\ref{appendix:QA-generation}.
The number of constructed questions is shown in the \texttt{Const-Q} column of Table~\ref{tab:dataset}. 
To prevent any domain from dominating the evaluation, we randomly sample 500 questions for each setting (1a, 2a, 4a, and 6a) in each domain for evaluation, resulting in 2,000 QA pairs per domain. 

\subsection{Construction Characteristics}
Our construction pipeline ensures that: 1) all questions in MACE have clear, complete, and correct ground-truth answers; 2) questions with different numbers of valid answers are explicitly separated and sufficiently represented, enabling analysis of answer-count effects and realistic mixed-question scenarios; and 3) questions across varying answer-count settings are grounded in the same knowledge types and generated from the same prompt templates, differing only in factors such as year range, thereby maintaining comparable difficulty.

\section{Experimental Setup}
\label{sec:experiment_setup}
In this section, we describe our experimental settings, covering the categorization of confidence estimation methods, models, and evaluation. 

\subsection{Confidence Estimation Methods}
\label{setup:method}

We extensively evaluate 15 existing confidence estimation methods and broadly categorize them into single-turn and double-turn methods.

\paragraph{Single-turn.} This class of methods estimates confidence along with answer generation and can be further divided into two types.
\begin{itemize}[leftmargin=0.8em, itemsep=0pt, topsep=2pt]

\item \textbf{Question-level.} These methods quantify the entropy of the model's output distribution. {Prediction Entropy \textit{(Prob Entropy)}}~\citep{malinin2020uncertainty} directly measures the entropy of the answer probability space; {Length-Normalize Prediction Entropy\textit{(N-Prob Entropy)}}~\citep{kadavath2022language} refines this by normalizing for sequence length; and {Semantic Entropy \textit{(Sem Entropy)}}~\citep{kuhn2023semantic} focuses on the semantic meaning regardless of the specific formatting.
\item \textbf{Answer-level Confidence}. These methods evaluate the confidence on a specific answer. \textit{Verb} prompts the model to directly express confidence in natural language, while \textit{Verb-Topk}~\citep{jiang2021can} outputs confidence  for  top-$k$ candidates. 
{Consistency\textit{(Consis)}}~\citep{Manakul2023SelfCheckGPTZB} measures confidence via agreement across multiple sampled outputs. \textit{Consis-Verb} and \textit{Consis-Verb-Topk}~\citep{xiong2023can} further weight this consistency using the verbalized scores from \textit{Verb} and \textit{Verb-Topk}. \textit{Perplexity} derives confidence from the sequence's generation probability.
\end{itemize}

\paragraph{Double-turn.} 
These methods adopt a post-hoc approach, measuring confidence through a secondary query after the initial answer is produced.
\begin{itemize}[leftmargin=0.8em, itemsep=0pt, topsep=2pt]
  \item \textbf{$P_{\text{True}}$}~\citep{kadavath2022language}. The model verifies its answer via a binary \texttt{True}/\texttt{False} judgment. Specifically, \textit{$P_{\text{True}}\text{-Consis}$} approximates the confidence of \texttt{True} via black-box sampling, whereas \textit{$P_{\text{True}}\text{-Prob}$} directly extracts the token's probability. Their conditioned variants, \textit{$P_{\text{True}}\text{-Consis-Cand}$} and \textit{$P_{\text{True}}\text{-Prob-Cand}$}, augment the verification prompt with multiple candidate answers to assist the judgment.

\item \textbf{{Self-Ask}}~\citep{tian2023just}. The model explicitly verbalizes a numerical confidence score in natural language. \textit{Self-Ask} performs this assessment directly, while \textit{Self-Ask-Cand} incorporates multiple candidate answers as assistance.
\end{itemize}

\noindent Detailed formulations of each method are provided in Appendix~\S~\ref{appendix:Method}. Compared with double-turn methods, we speculate that single-turn methods could be more strongly influenced by the number of answers, as it is closely tied to the probability of generating those answers.

\subsection{Models}
We employ a diverse set of representative language models across different scales and series.  
For open-source models, we include {Qwen2.5-Instruct} series (7B, 14B, 32B, 72B)~\citep{Yang2024Qwen25TR}, {LLaMA3.1-Instruct} series (8B, 70B)~\citep{grattafiori2024llama}, and {DeepSeek-V3}~\citep{liu2024deepseek}.
Additionally, we include two closed-source models GPT-4o-mini and GPT-4o~\citep{hurst2024gpt}.

\subsection{Evaluation}

We use \textbf{Accuracy} to measure QA performance, where an answer is considered correct if it matches the ground-truth answer, and report \textbf{Confidence} to reflect each method’s prediction. Following previous research~\citep{kuhn2023semantic, ni2025annotation}, we use \textbf{AUROC} (Area Under the Receiver Operating Characteristic Curve)~\citep{hanley1982meaning} to measure confidence quality. This metric evaluates whether the confidence assigned to correctly answered samples is higher than that assigned to incorrectly answered samples.
More details can be seen in~\S~\ref{app:experiment-details}.
For each metric, we report the average across all domains. Per-domain results are provided in Appendix~\ref{app:domain-4gt}.

\section{Results and Analysis} \label{sec:results} 
In this section, we examine how existing confidence estimation methods behave in multi-answer settings. We first analyze how QA performance and estimated confidence vary on questions with more correct answers, and then evaluate calibration performance under mixed questions with different answer counts. Due to space constraints, we focus on LLaMA-3.1-70B-Instruct, which shows trends consistent with other models; full results are provided in Appendix~\S~\ref{appendix:all model result}.
\begin{table}[t]
\centering
\caption{\textbf{Confidence scores of LLaMA-3.1-70B-Instruct} under different counts of correct answers. ``1a” denotes questions with a single correct answer, and the remaining settings follow the same convention.}
\small
\begin{tabular}{lcccc}
\toprule
\textbf{Method} & \textbf{1a} &\textbf{2a} & \textbf{4a} & \textbf{6a} \\
\midrule
Accuracy & 48.0 & 55.4 & 59.9 & 61.7 \\
\midrule
\multicolumn{5}{l}{\textbf{Single-turn}} \\
\midrule
\textit{Question-level} & & & & \\
Prob Entropy                             & 71.6 & 70.0 & 66.6 & 65.6 \\
N-Prob Entropy                           & 91.5 & 90.9 & 89.9 & 89.7 \\
Sem Entropy                              & 60.8 & 57.9 & 49.7 & 45.2 \\
\midrule
\textit{Answer-level} & & & & \\
Verb    & 94.8 & 94.7 & 93.1 & 92.3 \\
Verb-Topk         & 81.9 & 73.9 & 62.1 & 56.9 \\
Consis                    & 51.3 & 48.9 & 40.2 & 35.9 \\
Consis-Verb & 50.3 & 47.9 & 40.5 & 36.4 \\
Consis-Verb-Topk  & 47.6 & 45.9 & 38.4 & 34.2 \\
Perplexity                             & 77.6 & 76.4 & 73.0 & 71.0 \\
\midrule
\multicolumn{5}{l}{\textbf{Double-turn}} \\
\midrule
P\textsubscript{True}-Consis            & 79.3 & 79.3 & 77.0 & 75.5 \\
P\textsubscript{True}-Prob         & 79.2 & 79.0 & 76.8 & 75.4 \\
P\textsubscript{True}-Consis-Cand          & 88.1 & 87.9 & 87.1 & 87.9 \\
P\textsubscript{True}-Prob-Cand       & 88.1 & 87.8 & 87.1 & 87.8 \\
Self-Ask          & 63.6 & 60.9 & 56.1 & 54.4 \\
Self-Ask-Cand        & 64.5 & 63.5 & 59.6 & 56.8 \\
\bottomrule
\end{tabular}
\label{tab:confidence_llama70b}
\end{table}

\begin{figure}[t]
  \centering
  \includegraphics[width=\linewidth]{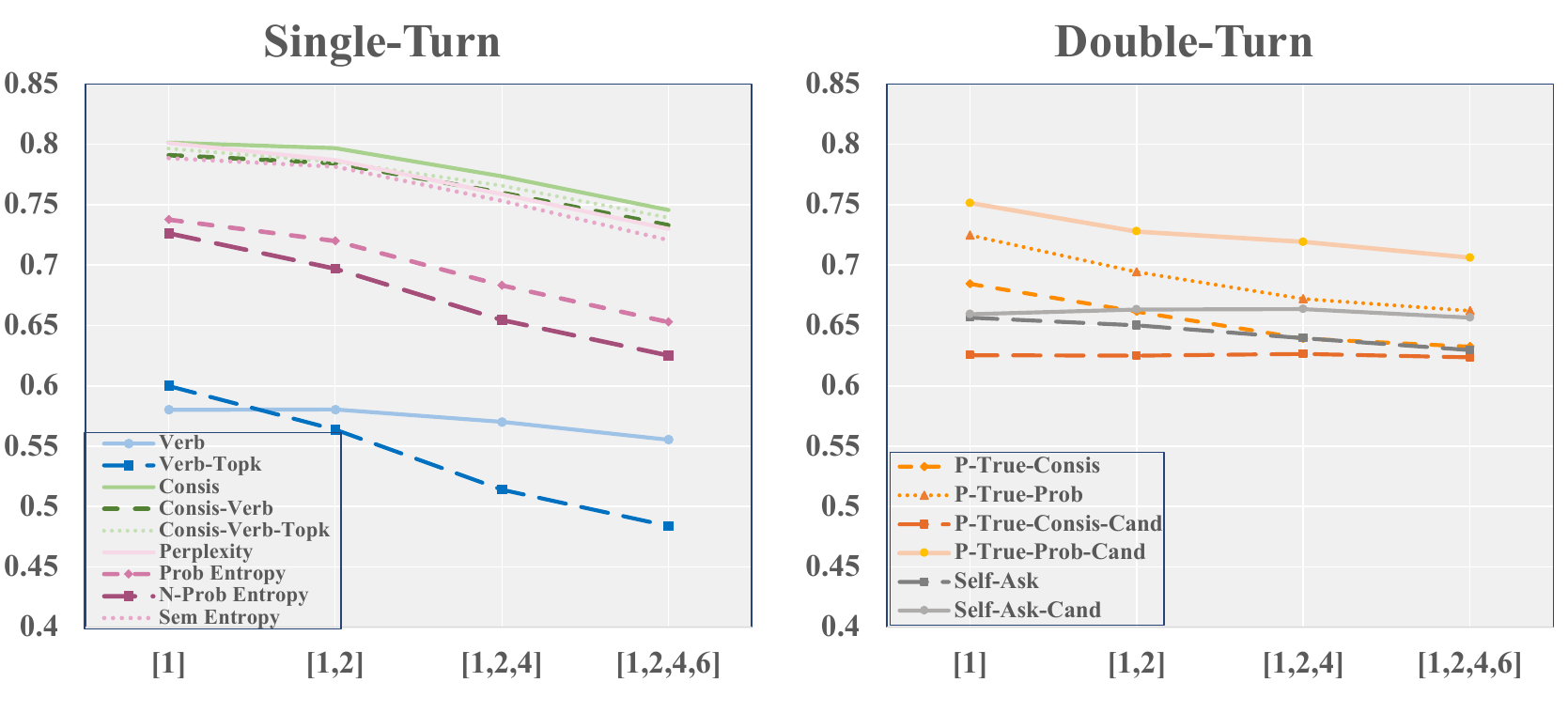}
\caption {\textbf{AUROC of LLaMA-3.1-70B-Instruct} on mixed questions with varying counts of correct answers.}
  \label{fig:auroc-llama-3p1-70b}
  \vspace{-0.5cm}
\end{figure}


\begin{figure}[t]
    \centering
    \includegraphics[width=0.4\textwidth]{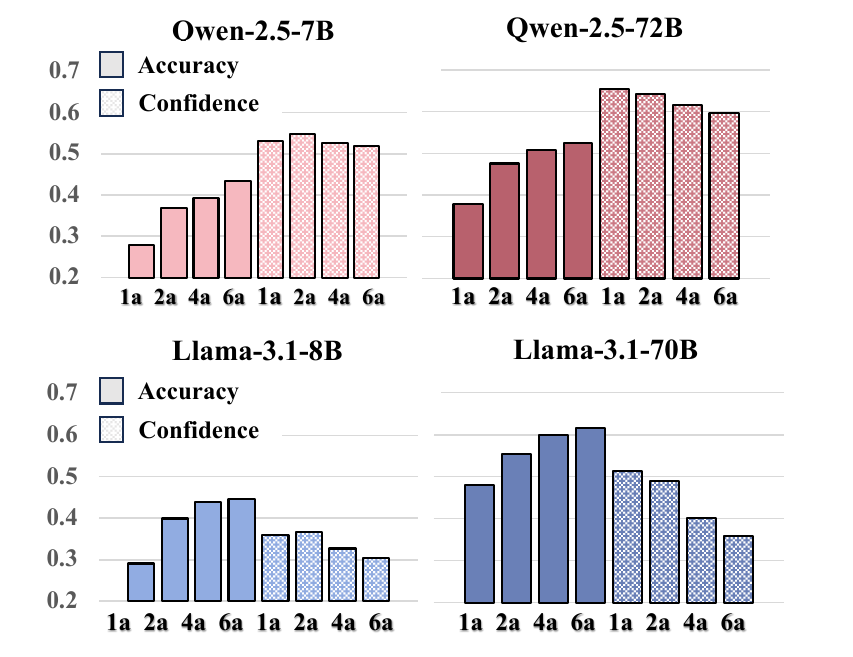}
    \caption{\textbf{QA performance and confidence variation} with the number of correct answers across model scales using \textit{Consistency} method.}
    \label{fig:model-scale-acc-conf}
\end{figure}


\subsection{Model Confidence Decreases as the Number of Correct Answers Increases} 
\label{sec:confidence change}

As demonstrated in Table~\ref{tab:confidence_llama70b}, increasing the number of ground-truth answers leads to a steady improvement in accuracy, suggesting that the questions become empirically easier. However, Table~\ref{tab:confidence_llama70b} reveals a consistent downward trend in confidence across most estimation methods. 
This indicates that miscalibration between confidence and QA performance arises in multi-answer settings for nearly all existing approaches.
Across different domains, the overall trend is similar. In more challenging domains, such as Region, the model’s confidence decreases less sharply (see Figure~\ref{fig:llama70B-avg-GT-Class-4gt}).


\paragraph{Larger models experience a sharper confidence drop.}
This phenomenon is shown in Figure~\ref{fig:model-scale-acc-conf}.
Within the \texttt{LLaMA} series, the 70B model exhibits a dramatic
confidence drop of \textbf{0.154} from the single-answer (1a) to
the six-answer (6a) setting, whereas the 8B model drops by only
\textbf{0.056}---nearly three times smaller.
We speculate that this is because larger models possess a much broader \answerspace. As the number of correct answers increases, larger models tend to generate a wider range of potential valid answers. In contrast, smaller models have limited knowledge coverage, so their generation space is often confined to a narrow subset of the possible answers. 
We provide a granular investigation into the impact of \answerspace, with detailed findings presented in Section ~\ref{sec:analysis_generation_space}. \looseness=-1

\subsection{Calibration Degrades on Mixed Questions with Varying Answer Counts}
To evaluate existing methods in a more realistic scenario where questions can have different numbers of valid answers, we mix the questions and examine overall calibration performance for each domain. AUROC results are illustrated in Figure~\ref{fig:auroc-llama-3p1-70b}. We observe a \textit{consistent downward trend in AUROC as questions become more heterogeneous in the number of correct answers for most of the methods}. This stems from the misalignment between rising accuracy and declining confidence estimates. \looseness=-1

\paragraph{Single-turn methods are more sensitive to the number of answers.}
Compared with double-turn methods, single-turn methods typically exhibit stronger calibration performance, except for verbalized approaches, due to models' inherent overconfidence~\citep{kadavath2022language}. However, their confidence scores are more sensitive to variations in the number of valid answers, leading to a faster degradation in calibration performance as the answer space diversifies. 
By contrast, double-turn methods are more stable across answer-count settings because they decouple confidence estimation from answer generation and rely on a post-hoc judge to verify answers; however, their overall calibration performance remains relatively limited. A detailed analysis is provided in~\S~\ref{app:method-analysis}.

\subsection{Why Does Confidence Decline More Sharply in Larger Models?}
\label{sec:analysis_generation_space}




We speculate that the decline in model confidence as the number of answers increases is related to the model's knowledge coverage. For models with limited knowledge, even as the number of correct answers increases, the candidate generation space may not expand substantially, leading to a relatively small confidence drop.
\paragraph {Definition of Knowledge Coverage.}
We consider a model's knowledge as the set of answers to which it assigns reasonable (i.e., not overly low) confidence. Following \cite{kuhn2023semantic}, we sample model responses and cluster semantically consistent answers to obtain the candidate generation space. We then filter out clusters with very low generation probabilities using a threshold of $\tau=0.1$, since such clusters may arise from randomness rather than the model's actual knowledge. We define knowledge coverage as the number of remaining high-confidence answer clusters. A detailed analysis of the cluster probability distribution and threshold selection is provided in Appendix~\S~\ref{appendix:ClusterProbDist}.


\begin{figure}[t]
    \centering
    \includegraphics[width=\linewidth]{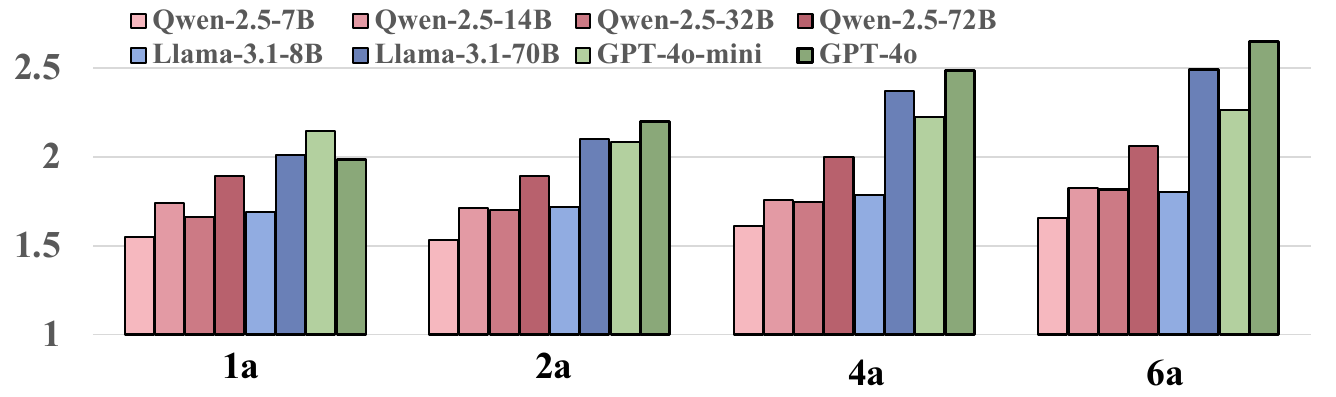}
\caption{
 \textbf{Average knowledge coverage across ground-truth (GT) set sizes. }Different color schemes denote different model families. Color intensity increases to indicate increasing model size.}
    \label{fig:model_scale_acc_conf_cluster}
\end{figure}

Figure~\ref{fig:model_scale_acc_conf_cluster} illustrates how \answerspace{} scales with model size, and Figure~\ref{fig:cluster_num_distribution} shows the distribution of cluster sizes. Larger models consistently exhibit a greater \answerspace{} than smaller ones, with the gap widening as the number of ground-truth answers increases. This suggests that larger models encode broader knowledge, whereas smaller models may fail to produce effective answers even when more correct answers exist.

Consequently, as the number of correct answers grows, larger models—by virtue of their broader knowledge—tend to produce a more diverse set of plausible responses, which dilutes response consistency and thereby lowers estimated confidence. Smaller models, by contrast, draw from a narrower candidate pool, resulting in a more gradual confidence decline. This disparity arises because knowledge coverage expands more rapidly in larger models, amplifying the confidence drop. We further validate this relationship quantitatively in Appendix~\S~\ref{app:Spearman Validity}, where Spearman correlation analysis confirms a strong association between knowledge coverage and confidence degradation.

\begin{figure}[h]
    \centering
    \includegraphics[width=1\linewidth]{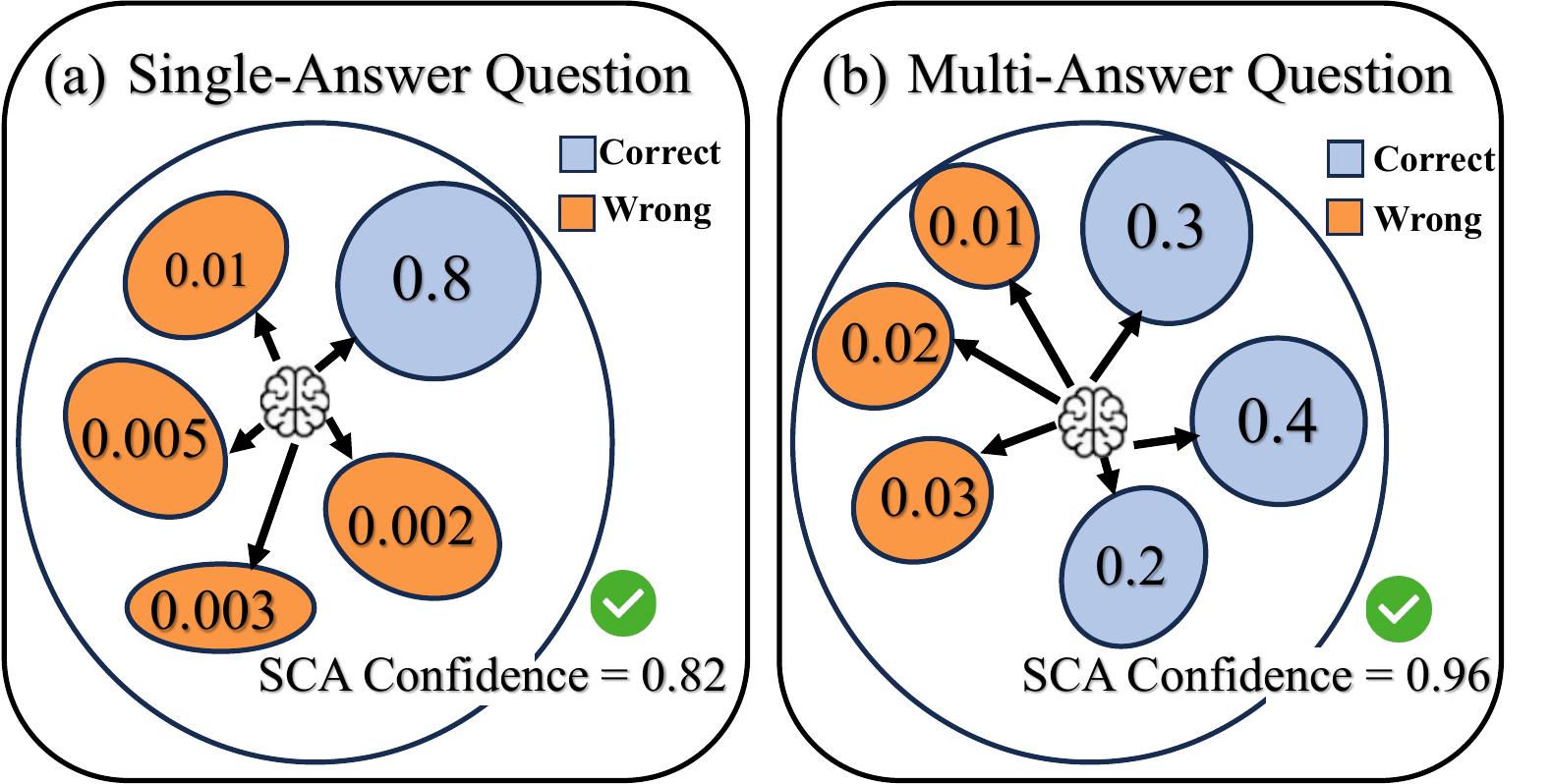}
\caption{ \textbf{Illustration of {SCA} method  without tuning.} 
As shown, confidence estimation is dominated by high-confidence clusters. SCA sums probabilities across all clusters, achieving consistent strong calibration performance in both single-answer and multi-answer settings.
}
\label{fig:cluster_motivation}
\vspace{-0.5cm}
\end{figure}

\section{Semantic Confidence Aggregation}
\label{sec:method}

In this section, we propose Semantic Confidence Aggregation (SCA), a method for confidence calibration in realistic multi-answer scenarios. The key idea is that overall confidence should reflect the aggregate confidence of all valid answers rather than rely on the single most confident response, and that response inconsistency (e.g., high entropy) does not necessarily imply low confidence.

\begin{table}[t]
\centering
\small
\setlength{\tabcolsep}{4pt}
\renewcommand{\arraystretch}{0.95}
\caption{
\textbf{AUROC score on {LLaMA-3.1-70B-Instruct}} of different calibration methods under increasing answer mixture.
$\tau=0$ denotes no-filtering, while $\tau\neq0$ uses the threshold value  tuned on the development set.
\textbf{Bold} marks the best, \underline{underlined} the second-best in each column.
$\dagger$ indicates the difference from our SCA method is not statistically significant ($p > 0.05$).
}
\begin{tabular}{lcccc}
\toprule
\small 
\textbf{Method} & \textbf{[1]} & \textbf{[1,2]} & \textbf{[1,2,4]} & \textbf{[1,2,4,6]} \\
\midrule
\multicolumn{5}{l}{\textbf{Single-turn}} \\
\midrule
\textit{Question-level} & & & & \\
Prob Entropy                                    & 74.5 & 72.6 & 69.0 & 66.1 \\
N-Prob Entropy                                  & 73.8 & 70.4 & 66.2 & 63.3 \\
Sem Entropy                                     & 78.9 & 78.6 & 75.8 & 72.5 \\
\midrule
\textit{Answer-level} & & & & \\
Verb                                            & 58.4 & 58.4 & 57.2 & 55.8 \\
Verb-Topk                                       & 60.7 & 56.2 & 51.2 & 48.4 \\
Consis                                          & 80.2$^\dagger$ & \underline{80.3} & 78.1 & \underline{75.3} \\
Consis-Verb                                     & 79.7$^\dagger$ & 79.0 & 76.6 & 73.8 \\
Consis-Verb-Topk                                & 80.0$^\dagger$ & 79.0 & 77.0 & 74.3 \\
Perplexity                                      & \textbf{80.9}$^\dagger$ & 79.4 & 76.5 & 73.8 \\

\midrule
\multicolumn{5}{l}{\textbf{Double-turn}} \\
\midrule
P\textsubscript{True}-Consis                    & 68.1 & 66.4 & 64.1 & 63.5 \\
P\textsubscript{True}-Prob                      & 72.9 & 70.1 & 67.6 & 66.7 \\
P\textsubscript{True}-Consis-Cand               & 62.6 & 62.7 & 62.9 & 62.5 \\
P\textsubscript{True}-Prob-Cand                 & 76.0 & 73.5 & 72.5 & 71.1 \\
Self-Ask                                        & 66.5 & 66.1 & 65.0 & 63.9 \\
Self-Ask-Cand                                   & 66.5 & 66.8 & 66.9 & 66.1 \\
\midrule
\multicolumn{5}{l}{\textbf{Confidence Aggregation Baselines}} \\
\midrule
{SNCA ($\tau=0.5$)}                              & 77.9 & 77.3 & 74.2 & 70.5 \\
{SNCA ($\tau=0$)}                                & 50.0 & 50.0 & 50.0 & 50.0 \\
{SFCA ($\tau=0.25$)}                             & 77.1 & 77.2 & 75.2 & 73.0 \\
{SFCA ($\tau=0$)}                                & 50.0 & 50.0 & 50.0 & 50.0 \\
\midrule
\multicolumn{5}{l}{\textbf{Ours}} \\
\midrule
{SCA ($\tau=0.05$)}                             & \underline{80.6} & \textbf{81.2} & \textbf{79.2} & \textbf{76.7} \\
{SCA ($\tau=0$)}                                & 80.5 & \textbf{81.2} & \underline{79.1} & \textbf{76.7} \\

\bottomrule
\end{tabular}
\label{tab:multi-answer-auroc-llama70b}
\end{table}



\subsection{Methodology}
Similar to Sem Entropy~\citep{kuhn2023semantic}, we begin by sampling multiple answers and clustering them based on semantic similarity.
Let $\mathcal{S} = \{s_1, \dots, s_N\}$ denote $N$ sampled responses for a given question.  We partition $\mathcal{S}$ into $M$ clusters $\mathcal{C}=\{C_1, \dots, C_M\}$. 
Following Sem Entropy, for a cluster $C_m$, we compute its confidence as the sum of the token-level probabilities of all responses within the cluster. The formulation is:
\begin{equation}
    P_{\text{SCA}}(C_m) = \sum_{s \in C_m} p(s),
\end{equation}
where $p(s)$ represents the token-level generation probability of sequence $s$.

Since low-probability responses may be generated by chance rather than representing answers the model considers likely correct, we sum only the probabilities of clusters whose confidence exceeds a threshold $\tau$. The set of valid clusters is defined as $\mathcal{C}_{\text{valid}} = \{ C_m \in \mathcal{C} \mid P(C_m) > \tau \}$ and the final confidence score for the question is computed as:
\begin{equation}
    \text{Conf}_{\text{SCA}} = \sum_{C_m \in \mathcal{C}_{\text{valid}}} P(C_m).
\end{equation}
Experimental results show that SCA achieves comparable performance at $\tau = 0$ to that obtained with the optimal $\tau$ (Table~\ref{tab:multi-answer-auroc-llama70b}). \textit{This indicates that simply summing token-level probabilities over all sampled responses is sufficient, without requiring clustering or filtering, thereby reducing computational cost}.
\looseness=-1

\subsection{Experimental Setup}
\label{sec:maps(exp setup)}
\paragraph{Additional Aggregation Methods.}
To verify whether using token-level probabilities to compute cluster confidence is appropriate, we compare two additional methods. These methods differ only in how cluster confidence is calculated.
 
\begin{itemize}[leftmargin=0.8em, itemsep=0pt, topsep=2pt]
    \item {SNCA (\textbf{S}emantic \textbf{N}ormalized  \textbf{C}onfidence \textbf{A}ggregation): } normalizes the aggregated token-level probabilities across all clusters to capture relative magnitudes:
$P_{\text{SNCA}}(C_m) = \frac{P_{\text{SCA}}(C_m)}{\sum_{j=1}^{M} P_{\text{SCA}}(C_j)}.$

    \item {SFCA (\textbf{S}emantic \textbf{F}requency  \textbf{C}onfidence \textbf{A}ggregation):}  defines probability of a cluster based on the count of samples within it without considering token-level probability: $P_{\text{SFCA}}(C_m) = \frac{|C_m|}{N}$.

\end{itemize}

\paragraph{Models and Datasets.}
Due to space constraints, experiments in this section focus on LLaMA-3.1-70B-Instruct; results for other models, which yield consistent conclusions, are reported in Appendix~\S~\ref{app:SCA auroc generalization}. We randomly sample 20\% of MACE from each domain as a development set and use the remaining data for testing.
For each aggregation method, two settings are evaluated: a fixed threshold $\tau=0$ and an optimal threshold tuned on the development set with statistical significance tested.



\begin{figure}
    \centering
    \includegraphics[width=1\linewidth]{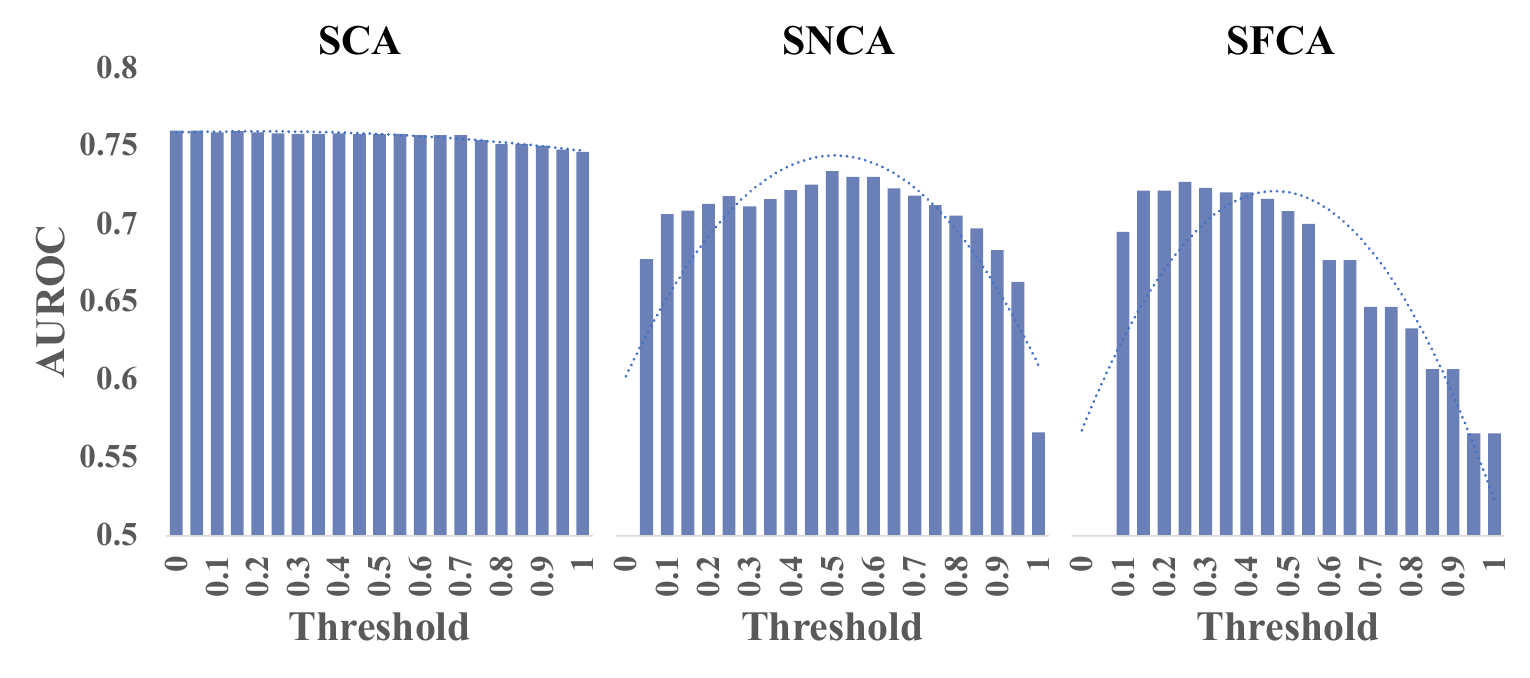}
   \caption{\textbf{Sensitivity of average AUROC across answer settings }to the threshold for aggregation methods.}
    \label{fig:sncs_secs}
\end{figure}
\subsection{Results and Analysis}
Table~\ref{tab:multi-answer-auroc-llama70b} shows calibration performance on the test set, and Figure~\ref{fig:sncs_secs} shows how the three aggregation methods vary with the threshold on the development set.
We observe that \textit{SCA matches the best-performing existing methods in the single-answer setting and achieves the strongest results across all three mixed multi-answer scenarios with $\tau=0$}. This means that SCA achieves strong calibration performance even without clustering or any filtering, simply by summing the token-level probabilities of all responses. Compared with SNCA and SFCA, SCA is less sensitive to the threshold. This indicates that the absolute values of token-level probabilities inherently capture the model’s perception of answer correctness, assigning high probabilities to likely correct answers and low probabilities to spurious ones, as shown in Figure~\ref{fig:cluster_motivation}.
Additionally, we evaluate on the Natural Questions~\citep{kwiatkowski2019natural} dataset to test generalization; as shown in Table~\ref{tab:nq-results}, SCA also achieves competitive performance on this single-answer benchmark.

\section{Conclusion}
\label{sec:conclusion}
This work extends confidence calibration from single-answer QA to a more general setting with mixed numbers of ground-truth answers. We introduce the MACE benchmark and evaluate existing calibration methods and find that, as the number of correct answers increases, LLMs achieve higher QA accuracy but lower estimated confidence, leading to severe miscalibration. Consequently, nearly all these methods collapse in realistic settings with mixed answer cardinalities. To address this, we propose SCA, which aggregates probabilities of multiple high-confidence answers rather than relying on the top one, achieving state-of-the-art calibration in this general multi-answer setting while preserving strong calibration on
single-answer questions.

\clearpage
\section*{Limitations}
\label{sec:limitations}
While this study offers novel insights into confidence estimation
within multi-answer settings, several limitations warrant further
consideration.
First, our benchmark employs formulaic prompt templates
(``Name one X that satisfies Y'') to ensure intra-domain difficulty
homogeneity and controlled answer scaling
(Appendix~\ref{appendix:QA-generation}).
Although this design is a deliberate methodological choice to isolate
confidence–rank dynamics, it sacrifices naturalistic diversity in
question phrasing.
Extending the benchmark with more heterogeneous surface forms---while
preserving difficulty control---is an important direction for future work.
Second, our evaluation is currently restricted to English-language
datasets.
Extending this analysis to multilingual or cross-lingual contexts
remains essential for a more comprehensive understanding of calibration
behavior across languages.
Third, our scope primarily encompasses short factual responses.
Since real-world applications often involve open-ended and compositional
generation, investigating confidence behavior in long-form contexts is a
crucial next step.
Nonetheless, we believe our controlled evaluation framework provides a
useful foundation for developing more robust calibration methods in
realistic multi-answer scenarios.

\section*{Ethical Considerations}

All data used in this study are derived from publicly available and legally accessible sources.  
We employ both open-weight and API-based large language models under their respective usage policies.  
All experiments are conducted strictly for academic research purposes, without involving human subjects or sensitive content.


\bibliography{acl/reference}

\appendix

\begin{table*}[t]
\centering
\small
\caption{Comparison of existing benchmarks with respect to three
properties required for diagnosing \emph{probability mass split}
in multi-answer factual QA.
\textbf{P1}: exhaustive ground truth;
\textbf{P2}: multi-answer questions;
\textbf{P3}: controlled answer cardinality.
Categories are organized by \emph{answer-space structure}.}
\label{tab:benchmark-gap}
\renewcommand{\arraystretch}{1.25}
\begin{tabular}{@{} l l c c c p{5.6cm} @{}}
\toprule
\textbf{Category} & \textbf{Representative} & \textbf{P1} & \textbf{P2} & \textbf{P3} & \textbf{Key Limitation} \\
\midrule
\multicolumn{6}{@{}l}{\textit{Closed answer space (fixed option sets)}} \\[2pt]
Single-correct MC
  & MMLU-PRO, ARC
  & \cmark & \xmark & \xmark
  & Fixed option sets bypass open-ended probability dispersion
    over the full vocabulary. \\
Multi-correct MC
  & SATA-Bench
  & \cmark & \cmark & \xmark
  & Correct-option count is not systematically controlled,
    precluding cardinality analysis. \\
\midrule
\multicolumn{6}{@{}l}{\textit{Open answer space (free-form generation)}} \\[2pt]
Single-answer
  & TriviaQA, NQ, Mintaka
  & \cmark & \xmark & \xmark
  & Single-entity answers; exhaustiveness is trivially satisfied
    but no mass split can occur. \\
Multi-answer (non-exhaustive)
  & WebQuestions, CWQ
  & \xmark & \cmark & \xmark
  & Incomplete knowledge bases yield non-exhaustive ground truth,
    contaminating calibration measurement. \\
\midrule
\textbf{Multi-answer (controlled)}
  & \textbf{MACE (Ours)}
  & \cmark & \cmark & \cmark
  & --- \\
\bottomrule
\end{tabular}
\end{table*}

\section{Limitations of Existing Calibration Benchmarks}
\label{sec:limitation-benchmarks}

A well-calibrated language model should assign confidence scores
that faithfully reflect its actual likelihood of being correct.
Measuring this property in multi-answer factual QA, however,
places stringent demands on the evaluation benchmark.
In particular, we argue that three properties must be
\emph{jointly} satisfied (see Table~\ref{tab:benchmark-gap}
for a summary):

\begin{enumerate}[leftmargin=*, nosep, label=\textbf{P\arabic*.}]
\item \textbf{Exhaustive Ground Truth.}
      For every question, the ground-truth answer set must enumerate
      \emph{all and only} the correct answers.  Without exhaustiveness,
      a model that generates a valid but unlisted answer will be
      wrongly penalized as hallucinating, corrupting any calibration
      measurement.
\item \textbf{Multi-Answer Questions.}
      The benchmark must contain questions whose correct answer set
      has cardinality $|\mathcal{A}| \geq 2$.
      Probability mass split---the phenomenon whereby a model's
      confidence in each individual correct answer is diluted because
      probability mass is dispersed across multiple valid
      completions---can only appear when multiple correct answers
      coexist.  Single-answer benchmarks are structurally blind to
      this effect.
\item \textbf{Controlled Answer Cardinality.}
      Questions should be organized into systematic cardinality tiers
      (e.g.\ $|\mathcal{A}| \in \{1, 2, 4, 6\}$) so that one
      can quantify how calibration degrades as the number of valid
      answers increases.  Without such control, any observed
      miscalibration cannot be attributed to mass split with
      confidence.
\end{enumerate}

We organize existing benchmarks along the dimension most relevant
to probability mass split---\emph{answer-space structure}---and
show that none satisfies all three properties simultaneously.
At the coarsest level, we distinguish benchmarks whose answer space
is \emph{closed} (a fixed set of options) from those requiring
\emph{open} free-form generation over the full vocabulary.
Within the open category, we further separate benchmarks by whether
correct answers are unique or multiple, and, in the latter case,
whether the ground truth is exhaustive.


\paragraph{Single-correct multiple choice.}
MMLU-PRO~\citep{wang2024mmlu} and ARC~\citep{clark2018think}
present a small, closed option set per question with exactly one
correct choice.
While this format enables clean calibration measurement, the model's probability is confined to a
predefined option set (e.g.\ A/B/C/D), entirely bypassing the
open-ended probability dispersion over the full vocabulary that
characterizes free-form generation.
They fail \textbf{P2} and \textbf{P3}.

\paragraph{Multi-correct multiple choice.}
SATA-Bench~\citep{xu2025sata} adopts a ``select all that apply'' format,
permitting multiple correct options per question
(\textbf{P1}~$\checkmark$, \textbf{P2}~$\checkmark$).
Nevertheless, the model still operates over a closed, small option set
rather than generating answers freely, limiting ecological validity
for studying open-ended probability dispersion.
More critically, the number of correct options per question is not
systematically controlled but varies arbitrarily, making it impossible
to trace calibration degradation as a function of answer cardinality.
This fails \textbf{P3}.


\paragraph{Open single-answer QA.}
Benchmarks such as TriviaQA~\citep{joshi2017triviaqa},
Natural Questions~\citep{kwiatkowski2019natural},
and Mintaka~\citep{sen2022mintaka} target factoid questions
with a single correct entity
(e.g.\ \emph{``Who wrote Hamlet?''} $\rightarrow$
\emph{Shakespeare}).
Because $|\mathcal{A}|=1$ by construction, exhaustiveness is
trivially satisfied (\textbf{P1}~$\checkmark$), yet probability
mass split cannot occur, failing both \textbf{P2} and \textbf{P3}.
These benchmarks are therefore suitable for studying single-answer
calibration but structurally incapable of diagnosing the multi-answer
dispersion phenomenon.

\paragraph{Open multi-answer QA (non-exhaustive).}
WebQuestions~\citep{berant2013semantic} and
ComplexWebQuestions~\citep{talmor2018web} derive answers from
structured knowledge bases, naturally yielding some multi-entity
answer sets (\textbf{P2}~$\checkmark$).
However, knowledge bases suffer from systematic incompleteness:
entities and relations are frequently missing.
As a result, the ground-truth answer sets are \emph{not}
exhaustive---a model may produce a factually correct answer that
the knowledge base simply does not contain---violating \textbf{P1}.
This incompleteness directly contaminates calibration evaluation,
as correct predictions are miscounted as errors.
Moreover, answer cardinality is an uncontrolled element of the
structure, failing \textbf{P3}.

\medskip
\noindent
In summary, the landscape of existing benchmarks leaves a critical
gap: \emph{no current dataset simultaneously provides exhaustive
ground truth, multi-answer questions, and controlled answer
cardinality.}
Closed-format benchmarks (single- and multi-correct MC) sidestep
open-ended probability dispersion altogether; open single-answer
benchmarks are structurally blind to mass split; and the only open
multi-answer benchmarks rely on incomplete knowledge bases that
corrupt calibration measurement.
This gap makes it impossible to systematically study how probability
mass split affects model calibration as the number of valid answers
scales.
To fill it, we introduce \textbf{MACE}, the first benchmark
designed to satisfy \textbf{P1}--\textbf{P3} jointly, enabling
controlled, fine-grained analysis of calibration under multi-answer
probability dispersion.

\section{The MACE Benchmark}
\label{appendix:dataset-construction}
\subsection{Domain Identification}
\label{app:domain-identification}

Selecting appropriate factual domains is a prerequisite for a
reliable and scalable benchmark.
We begin by articulating three guiding principles:

\begin{enumerate}[leftmargin=*]
\item \textbf{Answer Verifiability.}
      Every question must admit a \emph{fixed-size} set of correct answers
      that can be verified objectively (i.e.\ non-subjectively).
      This rules out domains whose ground truth is opinion-dependent,
      time-sensitive with high update frequency, or otherwise ambiguous.

\item \textbf{Diversity under Verifiability.}
      Subject to the verifiability constraint, the chosen domains should
      collectively span heterogeneous knowledge categories so that the
      benchmark reflects a broad spectrum of real-world information needs
      rather than a single topical cluster.

\item \textbf{Large-Scale Data Availability.}
      Each domain must contain a sufficiently large pool of candidate
      questions (on the order of thousands) so that we can freely control
      the number of constraints per query and the overall evaluation cost,
      rather than being bottlenecked by scarce data.
\end{enumerate}

\newtcolorbox{promptbox}[2][]{%
  enhanced,
  breakable,
  colback      = gray!8,
  colframe     = black!80,
  coltitle     = white,
  colbacktitle = black!85,
  fonttitle    = \small\bfseries,
  title        = {#2},
  boxrule      = 0.6pt,
  arc          = 1.5pt,
  left=8pt, right=8pt, top=6pt, bottom=6pt,
  fontupper    = \small,
  before upper = {\setlength{\parskip}{4pt}\setlength{\parindent}{0pt}},
  #1  
}

\begin{promptbox}[float=t, label={box:domain-brainstorm}]{Prompt for LLM-Assisted Domain Brainstorming}

\textbf{System Prompt:}

You are an expert research assistant helping to design a factual-QA
benchmark. Your task is to brainstorm knowledge domains that satisfy
a set of strict selection criteria provided by the user. You should
propose candidate domains, justify each one, and also provide
counter-examples of domains that appear suitable but actually fail
the criteria.

\tcblower

\textbf{User Prompt:}

\texttt{\#\# Instruction}

I am constructing a factual-QA benchmark. Each question asks the
model to produce a \emph{complete set} of entities that satisfy
several constraints simultaneously. I need to choose knowledge
domains that satisfy \textbf{all} of the following criteria:

\begin{enumerate}[leftmargin=1.5em, nosep, label=\arabic*.]
\item \textbf{Objective \& fixed-size answers.}
      The ground-truth answer set is uniquely determined and easily
      verifiable against public sources (e.g.\ Wikipedia, official
      records). The number of correct answers is finite and
      well-defined.
\item \textbf{Diversity.}
      The domains should span different areas of human knowledge
      (science, geography, politics, culture, etc.).
\item \textbf{Scalability.}
      Each domain should support thousands of questions with varied
      constraint combinations, so that the benchmark can be scaled
      up cheaply.
\end{enumerate}

Please suggest 15--20 candidate domains with a brief justification
for each. \textbf{Also provide counter-examples}---domains that
\emph{seem} suitable but actually violate one or more criteria---with
an explanation of why they fail.

\noindent\rule{\linewidth}{0.3pt}

\texttt{\#\# Counter-Example Guidance}

Below are several examples we provide to guide your reasoning.
Each illustrates a domain that initially appears suitable but
actually violates one or more of the above criteria:

$\bullet$\ ``Best-selling novels of a given decade''---answer sets are
publisher-dependent and vary across ranking sources, violating
\textbf{Criterion~1} (verifiability).

$\bullet$\ ``Famous paintings in a particular art movement''---the
membership of an art movement is subjective; experts disagree on
boundary cases, violating \textbf{Criterion~1}.

$\bullet$\ ``Endangered species in a region''---the IUCN Red List is
updated continuously, so the answer set is temporally unstable,
making large-scale verification expensive and violating
\textbf{Criterion~1~\&~3}.

$\bullet$\ ``Chemical elements with a specific property''---while
verifiable, the total number of elements is only 118, severely
limiting constraint diversity and violating \textbf{Criterion~3}.

$\bullet$\ ``Olympic host cities''---verifiable and objective, but the
total pool is small ($<$40 entries), making it hard to construct
thousands of distinct questions, violating \textbf{Criterion~3}.

\bigskip
Based on the criteria and counter-examples above, propose your
candidate domains.

\end{promptbox}

\paragraph{LLM-assisted brainstorming.}
Guided by these principles, we conduct iterative brainstorming sessions
with large language models (specifically GPT-4o) to enumerate candidate
domains.

After multiple rounds of LLM proposals and manual expert deliberation,
we converge on \textbf{six} representative domains that collectively
satisfy all three criteria.
Crucially, the ground-truth answers for every domain can be
objectively verified against Wikipedia, ensuring reproducibility
and low annotation cost:

\begin{itemize}[leftmargin=*]
\item \domainAward~(\Award): winners of well-documented prizes
      (e.g.\ Nobel, Academy Awards), whose complete recipient lists
      are maintained on Wikipedia with authoritative sourcing.
\item \domainOffice~(\Office): holders of political offices with
      clear start/end dates, systematically catalogued in
      Wikipedia's officeholder tables.
\item \domainCountry~(\Country): sovereign-state attributes such as
      organizational memberships and geographic properties,
      recorded in structured Wikipedia infoboxes.
\item \domainMath~(\Math): number-theoretic or combinatorial
      questions with provably unique answer sets, verifiable
      via Wikipedia's mathematical reference articles (e.g.\ lists
      of primes, Fibonacci numbers).
\item \domainRiver~(\River): hydrological and geographic facts
      about rivers (e.g.\ length, tributaries, countries traversed),
      documented in Wikipedia with well-established reference data.
\item \domainLanguage~(\Language): official and national language
      designations, recorded in Wikipedia's country-level articles.
\end{itemize}


\paragraph{Limitations of LLM-assisted brainstorming.}
While LLM-driven enumeration accelerates the exploration of the domain
space, it introduces several limitations that must be acknowledged:

\begin{enumerate}[leftmargin=*,label=(\roman*)]
\item \textbf{Surface-level plausibility bias.}
      LLMs tend to propose domains that \emph{sound} factual (e.g.\
      ``famous inventions,'' ``influential scientists'') without
      rigorously checking whether the answer sets are truly objective.  Human verification is therefore indispensable.

\item \textbf{Popularity bias.}
      LLM suggestions are skewed toward domains that are
      well-represented in their training corpora (e.g.\ Western awards), potentially under-representing
      knowledge areas from less-documented cultures or disciplines.

\item \textbf{Hallucinated counter-examples.}
      When asked to provide domains that violate our criteria, LLMs
      occasionally fabricate plausible-sounding but incorrect
      justifications (e.g.\ claiming a domain is subjective when it is
      not).  All counter-examples were therefore independently verified
      by the authors.
\end{enumerate}

\noindent
To mitigate these issues, every candidate domain proposed by the LLM
underwent a two-stage human review: (1)~a pilot feasibility check in
which we attempted to construct at least 200 questions with verified
ground truth, and (2)~expert deliberation among the authors to confirm
diversity and real-world relevance.

\subsection{Triplet Collection \label{sec:triplet_collection}}

Following prior work on knowledge-grounded evaluation~\cite{sen2022mintaka}, we adopt Wikidata as our primary knowledge source for entity retrieval, owing to its broad coverage, structured schema, and community-maintained quality.

We represent each piece of knowledge as a formal triplet $(s, r, \mathcal{O})$, where:
(1)~the \textbf{subject} $s$ denotes the primary entity or anchor of the query (e.g., \emph{Nobel Prize in Physics});
(2)~the \textbf{relation} $r$ denotes the semantic predicate connecting the subject to its answers (e.g., \emph{laureate of});
(3)~the \textbf{object set} $\mathcal{O} = \{o_1, o_2, \dots, o_k\}$ contains all valid ground-truth entities satisfying relation $r$ for subject $s$.
This formalized triplet representation provides a unified, machine-readable structure that facilitates downstream processing, including automatic question generation, answer verification, and controlled cardinality manipulation.
By varying the cardinality of the object set $|\mathcal{O}| \in \{1, 2, 4, 6\}$, we systematically control the difficulty and multi-answer complexity of each instance.

Except for the \textit{Math} domain, which is generated through rule-based synthesis, all other domains are collected via the Wikidata SPARQL endpoint\footnote{\url{https://query.wikidata.org/sparql}}.
Specifically, for each domain, we query Wikidata to retrieve relevant subjects and their associated factual objects based on predefined relations (see Table~\ref{tab:domain-schema}).
For instance, in the \textit{Award} domain, we retrieve the entity \emph{Nobel Prize in Physics} as the subject and obtain its \emph{laureates} in a specific year as the objects, linked by the relation ``is officially awarded to.''
To ensure alignment with common model knowledge, we restrict the temporal coverage to the years 1800--2023.


\begin{table}[t]
\centering
\small
\setlength{\tabcolsep}{10pt}
\caption{
\textbf{Counts of subjects remaining after each filtering step.} \textbf{Raw}: Total count of retrieved subjects. \textbf{Pop.}: Count after popularity filter. \textbf{Val.}: Count after validity filter. \textbf{Man.}: Count after manual verification.
}
\label{tab:filter}
\begin{tabular}{lrrrr}
\toprule
\textbf{Domain} & \textbf{Raw} & \textbf{Pop.} & \textbf{Val.} & \textbf{Man.} \\
\midrule
\Award    & 8,422  & 4,168 & 206   & 175 \\
\Office   & 7,306  & 5,000 & 289   & 159 \\
\Country  & 183    & 183   & 129   & 99  \\
\River    & 10,357 & 3,726 & 910   & 781 \\
\Language & 7,251  & 3,830 & 1,012 & 847 \\
\bottomrule
\end{tabular}
\end{table}

\subsection{Heuristic-Based Filtering}
\label{appendix:dataset-filtering}

To ensure data quality, we apply a two-stage filtering pipeline: a \textit{Popularity Filter} followed by a \textit{Validity Filter}. The \textit{Popularity Filter} operates on the \emph{subjects} of the knowledge triplets, while the \textit{Validity Filter} is applied to both \emph{subjects} and \emph{objects}.

\paragraph{Popularity Filter.}
Raw Wikidata queries often return a long tail of obscure or niche entities---for example, minor regional awards with no public recognition, or little-known administrative subdivisions---that are unlikely to fall within the parametric knowledge of any language model. Including such entities would yield low-quality questions that models cannot reasonably be expected to answer, thereby introducing noise into the evaluation.
To address this, we rank all entity labels within each domain by their
12-month aggregate Wikipedia pageview counts, which serve as a proxy for
public familiarity. We apply two complementary retention criteria:
(i)~we retain any item whose pageview count exceeds 1{,}000, ensuring
that all entities with a meaningful level of public recognition are
preserved regardless of their intra-domain rank;
(ii)~within each domain, we further cap the candidate pool at the top
5{,}000 entries by pageview rank, keeping the dataset at a manageable
scale for subsequent human annotation and manual quality review.
Together, these criteria strike a balance between broad entity coverage
and practical annotation cost, ensuring that the surviving subjects are
both well-known enough to meaningfully probe model knowledge and
tractable for rigorous human evaluation.

\paragraph{Validity Filter.}
Even after popularity-based pruning, entities retrieved from the Wikidata API frequently contain artifacts of incomplete or inconsistent curation. Common issues include missing human-readable labels (where only a raw QID such as \texttt{Q12345} is returned), empty or placeholder fields, broken outgoing URLs, and unresolved entity references that point to deleted or merged items.
We therefore apply a \textit{Validity Filter} that removes any triplet whose subject or object exhibits one or more of the following defects: (i)~an absent or empty \texttt{rdfs:label} in the target language; (ii)~a Wikidata identifier that cannot be resolved to a valid entity page; (iii)~missing relational fields required by the domain schema (see Table~\ref{tab:domain-schema}).

\paragraph{Domain-Specific Refinement Criteria.}
Beyond these universal checks, we further impose domain-specific
refinement criteria. A key motivation is that Wikidata's coverage is
often \emph{incomplete}: for a given subject, only a subset of its true
objects may be recorded, which would introduce false negatives during
evaluation---a model's correct answer could be wrongly penalized simply
because it is absent from an incomplete ground-truth set. The
domain-specific rules below therefore serve as heuristic proxies for
\emph{completeness}: by retaining only entities whose Wikidata records
exhibit signs of thorough curation (e.g., continuous yearly entries,
sufficient object counts), we maximize the likelihood that the
surviving triplets capture the full set of valid answers.

\begin{itemize}[leftmargin=1.2em]
    \item \textbf{Award:} We retain only award subjects granted to a
    single laureate per year, with continuous yearly records and at
    least three objects per subject. The single-laureate-per-year
    constraint avoids partially recorded co-recipient lists, while
    requiring continuity and a minimum object count serves as evidence
    that the award's historical record on Wikidata is
    comprehensively maintained.

    \item \textbf{Office:} We preserve subjects held by exactly one
    person per year, requiring a minimum of ten recorded years per
    subject. Restricting to single-holder offices eliminates cases
    where Wikidata only logs a subset of concurrent holders, and the
    ten-year threshold filters out short-lived or sparsely documented
    positions whose records are likely incomplete.

    \item \textbf{Region:} We include only countries with at least
    three administrative subdivisions. Countries with fewer entries are
    often incompletely catalogued on Wikidata, and requiring a minimum
    count helps ensure the subdivision list is reasonably exhaustive.

    \item \textbf{River} \& \textbf{Language:} We keep entities
    associated with no more than six countries to avoid overly generic
    mappings (e.g., English is spoken in dozens of countries, making
    exhaustive enumeration impractical and verification unreliable).
    Capping at six increases confidence that the Wikidata object set is
    complete. To control annotation cost, we further sample 600
    instances from objects associated with a single subject.
\end{itemize}

\subsection{Manual Verification}
\label{sec:human_check}

Although our heuristic filters substantially improve data quality, 
automated pipelines alone cannot guarantee factual correctness. 
We therefore conduct a rigorous multi-stage human verification process 
to validate all harvested triplets before inclusion in the final benchmark.

\paragraph{Independent Dual Annotation.}
Two annotators with domain expertise independently review each domain's 
entries along three dimensions: 
(i)~\emph{factual correctness}---whether the subject--relation--object 
mapping reflects a true real-world fact 
(e.g., verifying that a listed laureate did receive the specified award 
in the stated year); 
(ii)~\emph{completeness}---whether the object set $\mathcal{O}$ contains 
all valid answers rather than a partial subset; 
and (iii)~\emph{label consistency}---whether the surface-form labels 
faithfully correspond to their underlying Wikidata entities, ruling out 
disambiguation errors or stale redirects. 
Each triplet is cross-checked against its supporting Wikidata metadata 
(e.g., temporal qualifiers, geographic linkage, and official entity descriptions) 
to confirm semantic validity.

\paragraph{Adjudication and Agreement.}
Disagreements between the two annotators are resolved through a 
structured adjudication session: both annotators present their evidence, 
and a consensus label is produced only when both parties agree or a 
third senior annotator is consulted for tie-breaking. 
We measure inter-annotator reliability using Cohen's $\kappa$, 
obtaining $\kappa = 0.94$ averaged across all domains, 
indicating near-perfect agreement and confirming that the annotation 
guidelines are well-defined and consistently applied.

\paragraph{Iterative Spot-Check.}
After the initial annotation pass, we perform iterative spot-checking 
to catch residual errors. In each round, we draw a stratified random 
sample of 50 instances per domain---balanced across different object-set 
cardinalities---and revalidate them for factual soundness, formatting 
consistency, and correct cardinality assignment. 
Any identified errors trigger a targeted re-review of all entries 
sharing the same relation or subject category. 
This verification--revision loop is repeated until a full spot-check 
round surfaces zero errors; in practice, convergence was reached after 
two complete rounds.

\begin{table*}[t]
\centering
\caption{\textbf{Detailed schema of relations and entity types across six factual domains.}
Each row specifies the semantic category of subjects and objects,
along with the predicate used to query the underlying knowledge base.}
\label{tab:domain-schema}
\small
\renewcommand{\arraystretch}{1.25}
\setlength{\tabcolsep}{10pt}
\begin{tabular}{@{} l l l l @{}}
\toprule
\textbf{Domain}
  & \textbf{Subject Type}
  & \textbf{Relation ($s \rightarrow o$)}
  & \textbf{Object Type} \\
\midrule
\textsc{Award}
  & Award Name
  & \emph{is officially awarded to}
  & Person \\[3pt]
\textsc{Office}
  & Political Position
  & \emph{is officially held by}
  & Person \\[3pt]
\textsc{Region}
  & Sovereign Country
  & \emph{contains as first-level division}
  & Administrative Division \\[3pt]
\textsc{Math}
  & Number Type
  & \emph{includes as an instance}
  & Number \\[3pt]
\textsc{River}
  & River Name
  & \emph{flows through geographically}
  & Country \\[3pt]
\textsc{Language}
  & Human Language
  & \emph{is spoken widely in}
  & Country \\
\bottomrule
\end{tabular}
\end{table*}

\begin{table*}[t]
\centering
\caption{\textbf{Core prompt templates for each factual domain.}
Parameterised slots (highlighted in \texttt{\{braces\}}) are dynamically
filled to control the number of ground-truth answers $k$.}
\label{tab:core_prompts}
\small
\renewcommand{\arraystretch}{1.35}
\setlength{\tabcolsep}{10pt}
\begin{tabular}{@{} l p{0.82\textwidth} @{}}
\toprule
\textbf{Domain} & \textbf{Prompt Template} \\
\midrule
\textsc{Award} &
\texttt{Name one person who received \textbf{\{award\}} between
\textbf{\{lower\_year\}} and \textbf{\{upper\_year\}}.} \\[4pt]

\textsc{Office} &
\texttt{Name one person who officially assumed the office of
\textbf{\{office\_name\}} between \textbf{\{lower\_year\}} and
\textbf{\{upper\_year\}}.} \\[4pt]

\textsc{Region} &
\texttt{Name one first-level administrative division of
\textbf{\{region\}} whose area lies between \textbf{\{lower\_area\}}
and \textbf{\{upper\_area\}}~km\textsuperscript{2}.} \\[4pt]

\textsc{Math} &
\texttt{Name one \textbf{\{number\_type\}} number between
\textbf{\{lower\_num\}} and \textbf{\{upper\_num\}}.} \\[4pt]

\textsc{Language} &
\texttt{Name one country where \textbf{\{language\_1\}},
\textbf{\{language\_2\}}, or \textbf{\{language\_3\}} was predominantly
spoken by native speakers.} \\[4pt]

\textsc{River} &
\texttt{Name one country through which the \textbf{\{river\_1\}},
\textbf{\{river\_2\}}, or \textbf{\{river\_3\}} flows.} \\
\bottomrule
\end{tabular}
\end{table*}

\subsection{QA Pair Generation}
\label{appendix:QA-generation}

\paragraph{Design Rationale for Prompt Templates.}
All six domains share a unified \textbf{``Name one X that satisfies Y''}
sentence frame (Table~\ref{tab:core_prompts}), where \texttt{X} specifies the
entity type (e.g., person, country, number) and \texttt{Y} encodes one or more
adjustable constraints (e.g., temporal range, geographic attribute, numeric
interval).
This deliberate design choice is motivated by our central evaluation goal:
measuring how a model's confidence distributes across \emph{successive} correct
answers when the underlying question difficulty is held constant.

Concretely, the controlled template structure offers three methodological
advantages:

\begin{itemize}[nosep,leftmargin=1.5em]
  \item \textbf{Intra-domain Difficulty Homogeneity.}
        Because all instances within a domain differ only in the
        \emph{parameterised slots} (e.g., \texttt{\{award\}},
        \texttt{\{lower\_year\}}--\texttt{\{upper\_year\}}),
        surface-level confounds---such as varying syntactic complexity,
        question length, or pragmatic implicature---are effectively eliminated.
        This ensures that when a model's confidence decreases from the first
        correct answer to the $k$-th, the drop can be attributed to the model's
        internal knowledge recall dynamics rather than to heterogeneous question
        difficulty.

  \item \textbf{Unambiguous Answer Scope.}
        The ``Name one \ldots'' prefix explicitly signals that the expected
        output is a single entity per response turn.
        This removes ambiguity in automated answer matching and avoids
        partial-credit edge cases that would arise with open-ended phrasings
        such as ``What are some \ldots'' or ``List \ldots''.

  \item \textbf{Controlled Multi-answer Scaling.}
        The parameterised constraints (temporal windows, numeric intervals,
        disjunctive entity lists) act as \emph{precision knobs}:
        widening a year range or adding an extra river to the disjunction
        increases the ground-truth answer count $k$ without altering
        the question's syntactic form or cognitive demand.
        This mechanism is the foundation of our answer-count control
        ($k \in \{1,2,4,6\}$).
\end{itemize}

\noindent
We acknowledge that formulaic templates sacrifice naturalistic diversity.
However, for a benchmark whose purpose is to isolate
\emph{confidence–rank dynamics under controlled difficulty},
eliminating surface variation is a deliberate and necessary methodological
decision.
We discuss extending the benchmark with more diverse question phrasings
while preserving difficulty control in Section~\ref{sec:limitations}.

\paragraph{Retrieval-based Domains.}
Each domain consists of a unique set of factual subjects, where each subject
is linked to one or more objects, forming foundational knowledge triplets
$(s, r, \mathcal{O})$.
To construct questions that elicit a specific number of correct answers
($k \in \{1, 2, 4, 6\}$), we dynamically adjust the parameterised constraints
within each template via a \textit{constraint relaxation} strategy,
modulating the relational scope so that the resulting ground-truth set
satisfies $|\mathcal{O}| = k$.
This is realised through two complementary approaches:

\begin{enumerate}[nosep,leftmargin=1.5em]
  \item \textbf{One-to-Many Expansion (Intra-subject)} ---
        applied in \textit{\Award}, \textit{\Office}, and \textit{\Country}.
        For subjects that naturally possess multiple objects under a broad
        relation, we relax the constraint parameters while keeping the subject
        fixed.
        For example, in the \textit{\Award} domain, narrowing the year window
        \texttt{\{lower\_year\}}--\texttt{\{upper\_year\}} to a single year
        typically yields $k\!=\!1$, whereas widening it to a five-year span
        can yield $k\!=\!4$ or $k\!=\!6$ verified laureates.
        Similarly, in \textit{\Country}, the area interval
        \texttt{\{lower\_area\}}--\texttt{\{upper\_area\}} is calibrated
        so that exactly $k$ first-level administrative divisions fall within the
        specified range.
        In all cases, the relaxation is tuned to produce \emph{exactly} $k$
        ground-truth objects.

  \item \textbf{One-to-Few Aggregation (Cross-subject)} ---
        applied in \textit{\Language} and \textit{\River}.
        When a single subject–relation pair cannot naturally yield $k$ objects,
        we construct multi-answer instances by combining multiple subjects that
        share the same relational type into a disjunctive query.
        As shown in Table~\ref{tab:core_prompts}, the \textit{\Language} template
        includes up to three languages
        (\texttt{\{language\_1\}}, \texttt{\{language\_2\}},
         \texttt{\{language\_3\}}),
        and the \textit{\River} template includes up to three rivers.
        The number of disjuncts and the specific entities are selected so that
        the union of their associated countries totals exactly $k$.
        During aggregation, we verify that the merged object sets are
        \emph{mutually exclusive} to prevent answer overlap.
        Since the combinatorial space of possible aggregations is vast,
        we cap the number of generated questions at 10{,}000 per domain.
\end{enumerate}

Together, the two paths ensure that every domain can be populated across
all four answer-count settings ($k \in \{1,2,4,6\}$) while each question
retains a well-defined scope and a uniform level of intrinsic difficulty.

\paragraph{Rule-based Domain.}
The \textit{\Math} domain is defined through five numeric category predicates
(Table~\ref{tab:number_types}), each acting as a pseudo-label
(e.g., prime numbers, perfect squares).
Because these categories span broad numerical ranges, constructing
questions around specific target numbers is unnecessary.
Instead, we generate queries by uniformly sampling integers from the interval
$[0,\;1{,}000{,}000]$.
For each sampled number $n$, the ground-truth label set $\mathcal{O}$ is
computed deterministically by evaluating which of the five predicates $n$
satisfies, and the numeric interval
\texttt{\{lower\_num\}}--\texttt{\{upper\_num\}} in the template is set to a
narrow window centred on $n$ that contains exactly $k$ qualifying numbers.
This provides wide coverage of numerical reasoning behaviours while
maintaining precise answer-count control.
Similarly, given the virtually unlimited pool of candidate integers,
we cap the total number of generated questions at 10{,}000.

\paragraph{Quality Control.}
All generated QA instances undergo a multi-stage quality pipeline:

\begin{enumerate}[nosep,leftmargin=1.5em]
  \item \textbf{Deduplication and Leakage Prevention.}
        We perform exact-match and near-duplicate removal
        (based on normalised question strings)
        and enforce that no answer entity appears in more than one domain,
        preventing cross-domain information leakage.

  \item \textbf{Clustering-based Stratified Sampling.}
        To construct a representative and balanced dataset,
        we encode all candidate instances per domain-$k$ partition
        with Sentence-BERT embeddings and apply K-Means clustering
        ($K\!=\!500$).
        From each cluster we select the instance nearest to its centroid,
        yielding 500 instances per answer-count setting.
        Across four settings ($k\!\in\!\{1,2,4,6\}$) this produces
        \textbf{2{,}000 distinct samples per domain}
        and a final dataset of \textbf{12{,}000 instances} over six domains.
\end{enumerate}

\begin{table}[t]
\centering
\caption{\textbf{Five mathematical number types} used to construct
questions in the \textsc{Math} domain.
Each defines a distinct numeric sequence within $[0,\,10^6]$.}
\label{tab:number_types}
\small
\setlength{\tabcolsep}{10pt}
\renewcommand{\arraystretch}{1.35}
\begin{tabular}{@{} l l @{}}
\toprule
\textbf{Number Type} & \textbf{Definition} \\
\midrule
\textsc{Prime}
  & $p>1,\; p\in\mathbb{N},\; \nexists\, d\in(1,p):\, d\mid p$ \\[3pt]
\textsc{Square}
  & $n=k^{2},\; k\in\mathbb{N}$ \\[3pt]
\textsc{Cube}
  & $n=k^{3},\; k\in\mathbb{N}$ \\[3pt]
\textsc{Fibonacci}
  & $F_{n}=F_{n-1}+F_{n-2},\; F_{1}=F_{2}=1$ \\[3pt]
\textsc{Triangular}
  & $T_{n}=\dfrac{n(n+1)}{2},\; n\in\mathbb{N}$ \\
\bottomrule
\end{tabular}
\end{table}

\section{Experiment Setup}

\subsection{Calibration Methods}
\label{appendix:Method}

\begin{itemize}[leftmargin=1.4em]

  \item \textbf{Prediction Entropy.}\cite{kadavath2022language}  
    To capture uncertainty across $M$ generated responses, we compute the mean token-level entropy over all samples:
    \begin{equation}
    U_{\text{Prob Entropy}} = -\frac{1}{M} \sum_{j=1}^{M} \sum_{i=1}^{|r_j|} p_{z_i} \log p_{z_i},
    \label{eq:pe}
    \end{equation}
    where $p_{z_i}$ represents the normalized probability distribution over tokens within the $j$-th generation.

    \item \textbf{Length-Normalized Prediction Entropy.}  \cite{malinin2020uncertainty}
    To avoid bias toward longer sequences, we normalize the entropy of each sampled response by its length:
    \begin{equation}
    U_{\text{N-Prob Entropy}} = -\frac{1}{M} \sum_{j=1}^{M} \frac{1}{|r_j|} \sum_{i=1}^{|r_j|} p_{z_i} \log p_{z_i}.
    \label{eq:lnpe}
    \end{equation}
    This normalization ensures comparable uncertainty estimates across outputs of varying lengths.

   \item \textbf{Semantic Entropy.}  \cite{kuhn2023semantic}
Given a set of semantic clusters $\{C_1,\dots,C_K\}$ obtained from multiple sampled responses, 
we compute a probability distribution over clusters by aggregating the sentence-level probabilities.  
Specifically, the probability assigned to cluster $C_k$ is
\begin{equation}
P(C_k) = \sum_{r_j \in C_k} P(r_j),
\label{eq:cluster_prob}
\end{equation}
where $P(r_j)$ denotes the generation probability of response $r_j$.
In standard semantic-entropy settings, these cluster probabilities are typically normalized to form a valid distribution:
\begin{equation}
\tilde{P}(C_k) = \frac{P(C_k)}{\sum_{i=1}^{K} P(C_i)},
\label{eq:cluster_prob_norm}
\end{equation}
yielding $\tilde{P}(C_k) \in [0,1]$ and $\sum_k \tilde{P}(C_k)=1$.

The semantic entropy is then computed as the Shannon entropy of the normalized cluster distribution:
\begin{equation}
U_{\text{SE}}
= - \sum_{k=1}^{K} \tilde{P}(C_k) \log \tilde{P}(C_k),
\label{eq:semantic_entropy}
\end{equation}
which quantifies the degree of semantic dispersion in the model's outputs—high values indicate diverse or inconsistent answer patterns, while low values correspond to concentrated semantic predictions.

\paragraph{From Entropy to Confidence.}
The three metrics above are measures of \emph{uncertainty}: higher values
indicate greater model hesitation.
To convert them into a unified \emph{confidence} score that is
monotonically increasing with model certainty and bounded in $(0,\,1]$,
we apply an exponential transformation:
\begin{equation}
\mathrm{Conf}(x) \;=\; e^{-H(x)},
\label{eq:entropy2conf}
\end{equation}
where $H(x) \in \{H_{\text{PE}},\, H_{\text{NPE}},\, H_{\text{SE}}\}$
denotes the entropy value for a given instance $x$.
When entropy is zero (the model is maximally certain), the confidence
equals~1; as entropy grows, confidence decays smoothly toward~0.
This transformation preserves the relative ordering of instances
and enables direct comparison with other confidence estimators
on a common scale throughout our analysis.

   \item \textbf{Verb.} \citep{jiang2021can}  
This method elicits confidence by prompting the model to verbalize a numeric confidence score alongside its answer.  
Given a question $x$, the model generates a textual response containing both the answer $\hat{y}$ and a confidence expression.  
The confidence value $C$ is then extracted from the generated text:
\begin{equation}
    C_{\text{Verb}}
    = \text{LLM}(x).
    \label{eq:vanilla_verb}
\end{equation}

    \begin{promptbox}[float=t, label={box:verbalized-single}]{Prompt Template for Verb}

\textbf{System Prompt:}

You are a knowledgeable assistant. When answering factual questions,
you must provide your answer along with a numeric confidence score
that reflects your genuine certainty about the answer being correct.

\tcblower

\textbf{User Prompt:}

\texttt{\#\# Instruction}

Please answer the following question and provide your confidence
as a numeric value between 0 and 1
(e.g., 0.85 means you are 85\% confident).
Use two-decimal precision for your confidence value.

\noindent\rule{\linewidth}{0.3pt}

\texttt{\#\# Input}

Question: \{question\}

\noindent\rule{\linewidth}{0.3pt}

\texttt{\#\# Output Format}

\begin{verbatim}
Answer: {answer}
Confidence: 
{confidence between 0 and 1
two-decimal precision}
\end{verbatim}

\end{promptbox}
    \vspace{6pt}

   \item \textbf{Verb\textsubscript{Topk}.} \citep{xiong2023can}  
This variant extends the verbalization setting by asking the model to provide $k$ distinct candidate answers  
$\{\hat{y}_1, \hat{y}_2, \dots, \hat{y}_k\}$ along with their respective confidences $\{C_1, C_2, \dots, C_k\}$.  
The final confidence is defined as the maximum confidence among semantically valid candidates:
\begin{equation}
    C_{\text{Verb\textsubscript{Topk}}} 
    = \max_{1 \le i \le k} \left( C_i \cdot \mathbb{I}\{\hat{y}_i \in \mathcal{Y}^*\} \right),
    \label{eq:topk_verb}
\end{equation}
where $\mathcal{Y}^*$ denotes the set of semantically valid answers.

    \begin{promptbox}[float=t, label={box:verbalized-prompt}]{Prompt Template for Verb\textsubscript{Topk}}

\textbf{System Prompt:}

You are a knowledgeable assistant. When answering factual questions,
you must provide your top-$k$ most likely answers along with a numeric
confidence score for each. Your confidence should reflect your genuine
certainty about each answer being correct.

\tcblower

\textbf{User Prompt:}

\texttt{\#\# Instruction}

Please list your top-$k$ most likely answers to the following question.
For each answer, report your numeric confidence between 0 and 1
(e.g., 0.85 means 85\% confident).
Use two-decimal precision for all confidence values.

\noindent\rule{\linewidth}{0.3pt}

\texttt{\#\# Input}

Question: \{question\}

\noindent\rule{\linewidth}{0.3pt}

\texttt{\#\# Output Format}

\begin{verbatim}
1. Answer: {answer_1}, 
Confidence: {confidence_1}
2. Answer: {answer_2}, 
Confidence: {confidence_2}
... up to k answers.
\end{verbatim}

\end{promptbox}

\end{itemize}

\begin{itemize}[leftmargin=1.5em, itemsep=0pt, topsep=2pt]
    \item \textbf{Consistency.} \cite{Manakul2023SelfCheckGPTZB} 
    {Consistency} score measures the proportion of model samples that agree with the aggregated prediction $\tilde{Y}$ among $M$ sampled outputs:
    \begin{equation}
    C_{\text{consistency}} = \frac{1}{M} \sum_{i=1}^{M} \mathbb{I}\{\hat{Y}_i = \tilde{Y}\},
    \label{eq:consistency}
    \end{equation}

    \item \textbf{Weighted Consistency}\cite{xiong2023can}
    To incorporate the self-reported confidence $C_i$ of each sampled answer, we further introduce a weighted variant:
    \begin{equation}
    C_{\text{conf}} =
    \frac{\sum_{i=1}^{M} \mathbb{I}\{\hat{Y}_i = \tilde{Y}\} \times C_i}
    {\sum_{i=1}^{M} C_i},
    \label{eq:weighted_consistency}
    \end{equation}
    where higher-confidence answers contribute proportionally more to the final score.  
    When $C_i$ is obtained from the \textit{Verb} or \textit{Verb\textsubscript{Topk}} methods, this variant corresponds to \textit{{Consistency-Verb}} and \textit{{Consistency-Verb-Topk}}, respectively.

\end{itemize}

    \begin{itemize}[leftmargin=1.5em, itemsep=0pt, topsep=2pt]
    \item \textbf{Perplexity.}\cite{Manakul2023SelfCheckGPTZB} Perplexity is computed as the average negative log-likelihood of token probabilities:
    \begin{equation}
    U_{\text{Perplexity}} = -\frac{1}{|r|} \sum_{i=1}^{N} \log p_{z_i},
    \label{eq:ppl}
    \end{equation}
    where $r$ denotes the generated sequence and $p_{z_i}$ is the model-assigned probability for the $i$-th token.

\end{itemize}

\begin{itemize}[leftmargin=1.5em, itemsep=0pt, topsep=2pt]

    \item \textbf{P\textsubscript{True}-Consis/P\textsubscript{True}-Prob} \citep{kadavath2022language}.  
    These methods estimate model confidence by prompting the LLM to explicitly verify whether its previously generated answer is correct.  
    The verification prompt $x'$ asks the model to respond with ``\texttt{True}'' or ``\texttt{False}'', and the resulting confidence is defined as the probability of the token ``\texttt{True}'':
    \begin{equation}
        C_{\text{P\textsubscript{True}}} = p(z_{\text{true}} \mid x').
        \label{eq:p_true}
    \end{equation}
    We implement two variants:  
    (1) In the \textbf{P\textsubscript{True}-Consis} variant, confidence is obtained by performing multiple sampling runs and taking the empirical frequency of the ``\texttt{True}'' response across samples;  
    (2) In the \textbf{P\textsubscript{True}-Prob} variant, we directly use the model’s internal probability (logit-derived softmax value) assigned to the ``\texttt{True}'' token, without repeated sampling.

   \begin{promptbox}[float=t, label={box:self-verify}]{Prompt Template for P\textsubscript{True}}

\textbf{System Prompt:}

You are a rigorous fact-checking assistant. Given a question and a
previously generated answer, you must judge whether the answer is
factually correct. Respond strictly with either \texttt{True} or
\texttt{False}---no explanations, no additional text.

\tcblower

\textbf{User Prompt:}

\texttt{\#\# Instruction}

You will be given a question and the model's previous answer.
Please respond with \texttt{True} if the answer is factually correct,
or \texttt{False} if it is incorrect.
Answer strictly with either ``True'' or ``False'' only.

\noindent\rule{\linewidth}{0.3pt}

\texttt{\#\# Input}

Question: \{question\} \\
Model answer: \{answer\}

\noindent\rule{\linewidth}{0.3pt}

\texttt{\#\# Output Format}

\begin{verbatim}
True / False
\end{verbatim}

\end{promptbox}

    \vspace{6pt}

    \item \textbf{P\textsubscript{True}-Consis-Cand/P\textsubscript{True}-Prob-Cand} \citep{kadavath2022language}.  
    To better handle multi-answer scenarios, we extend the verification prompt $x'$ by including a set of candidate answers $\{a_1, a_2, \dots, a_K\}$, allowing the model to evaluate its response in the context of multiple plausible answers:
    \begin{equation}
        C_{\text{P\textsubscript{True}-Consis-Cand}} = p(z_{\text{true}} \mid x', \{a_1, a_2, \dots, a_K\}).
        \label{eq:p_true_candidates}
    \end{equation}
    Similarly, two variants are considered:  
    (1) In the \textbf{P\textsubscript{True}-Consis-Cand} variant, confidence is computed as the empirical frequency of ``\texttt{True}'' responses across multiple sampling runs;  
    (2) In the \textbf{P\textsubscript{True}-Prob-Cand} variant, the probability assigned to the ``\texttt{True}'' token from the model’s logits is used directly as the confidence score.

   \begin{promptbox}[float=t, label={box:candidate-verify}]{Prompt Template for P\textsubscript{True}-Cand}

\textbf{System Prompt:}

You are a rigorous fact-checking assistant. Given a question, a
previously generated answer, and a list of candidate answers, you must
judge whether the model's answer is factually correct. Respond strictly
with either \texttt{True} or \texttt{False}---no explanations, no
additional text.

\tcblower

\textbf{User Prompt:}

\texttt{\#\# Instruction}

You will be given a question, the model's answer, and several candidate
answers. Please respond with \texttt{True} if the model's answer is
factually correct, or \texttt{False} if it is incorrect.
Answer strictly with either ``True'' or ``False'' only.

\noindent\rule{\linewidth}{0.3pt}

\texttt{\#\# Input}

Question: \{question\} \\
Model answer: \{answer\} \\
Candidate answers: \{candidate\_1\}, \{candidate\_2\}, \{candidate\_3\}, \ldots

\noindent\rule{\linewidth}{0.3pt}

\texttt{\#\# Output Format}

\begin{verbatim}
True / False
\end{verbatim}

\end{promptbox}
\end{itemize}

\begin{itemize}[leftmargin=1.5em, itemsep=0pt, topsep=2pt]

    \item \textbf{Self\textsubscript{Ask}.} \citep{tian2023just} 
    This method prompts the model to \textit{self-assess} and directly verbalize its own confidence score after producing an answer.  
    Given a question–answer pair $(x, \hat{y})$, the model is re-prompted with a meta-instruction asking it to output a numerical confidence value (e.g., “0.85”) that reflects how certain it is about $\hat{y}$.  
    The resulting confidence is defined as:
    \begin{equation}
        C_{\text{Self\textsubscript{Ask}}} = \text{LLM}(x', \hat{y}),
        \label{eq:selfask}
    \end{equation}
    where $x'$ denotes the self-assessment prompt and $\text{LLM}(\cdot)$ returns a scalar confidence value verbalized by the model.  

    \begin{promptbox}[float=t, label={box:post-hoc-conf}]{Prompt Template for Self\textsubscript{Ask}}

\textbf{System Prompt:}

You are a calibrated evaluator. After seeing a question and the
model's own answer, you must assess how confident the model should be
that its answer is correct. Output only a single numeric value between
0 and 1---no explanations, no additional text.

\tcblower

\textbf{User Prompt:}

\texttt{\#\# Input}

Question: \{question\} \\
Answer: \{answer\}

\noindent\rule{\linewidth}{0.3pt}

\texttt{\#\# Instruction}

How confident are you that the above answer is correct?
Please output only a number between 0 and 1
(e.g., 0.85 means 85\% confident).
Use two-decimal precision.

\noindent\rule{\linewidth}{0.3pt}

\texttt{\#\# Output Format}

\begin{verbatim}
{confidence between 0 and 1
two-decimal precision}
\end{verbatim}

\end{promptbox}

    \item \textbf{Self\textsubscript{Ask}-Cand.}  \cite{tian2023just}
    To better account for uncertainty among multiple plausible answers, this variant augments the verification prompt with a set of candidate answers $\{a_1, a_2, \dots, a_K\}$.  
    The model is instructed to assess its confidence in $\hat{y}$ given this broader context.  
    The confidence is computed as:
    \begin{equation}
        C_{\text{Self\textsubscript{Ask}-Cand}} = \text{LLM}(x', \hat{y}, \{a_1, a_2, \dots, a_K\}),
        \label{eq:selfask_c}
    \end{equation}
    where $x'$ includes both the original answer and the candidate list as context for self-evaluation.

  \begin{promptbox}[float=t, label={box:candidate-conf}]{Prompt Template for Self\textsubscript{Ask}-Cand}

\textbf{System Prompt:}

You are a calibrated evaluator. After seeing a question, the model's
own answer, and a set of candidate answers, you must assess how
confident the model should be that its original answer is correct.
Output only a single numeric value between 0 and 1---no explanations,
no additional text.

\tcblower

\textbf{User Prompt:}

\texttt{\#\# Input}

Question: \{question\} \\
Model answer: \{answer\} \\
Candidate answers: \{candidate\_1\}, \{candidate\_2\}, \{candidate\_3\}, \ldots

\noindent\rule{\linewidth}{0.3pt}

\texttt{\#\# Instruction}

Considering the candidate answers above, how confident are you that
the model's original answer is correct?
Please output only a number between 0 and 1
(e.g., 0.85 means 85\% confident).
Use two-decimal precision.

\noindent\rule{\linewidth}{0.3pt}

\texttt{\#\# Output Format}

\begin{verbatim}
{confidence between 0 and 1
two-decimal precision}
\end{verbatim}

\end{promptbox}

\end{itemize}

\begin{figure*}[t]
    \centering
    \includegraphics[width=\textwidth]{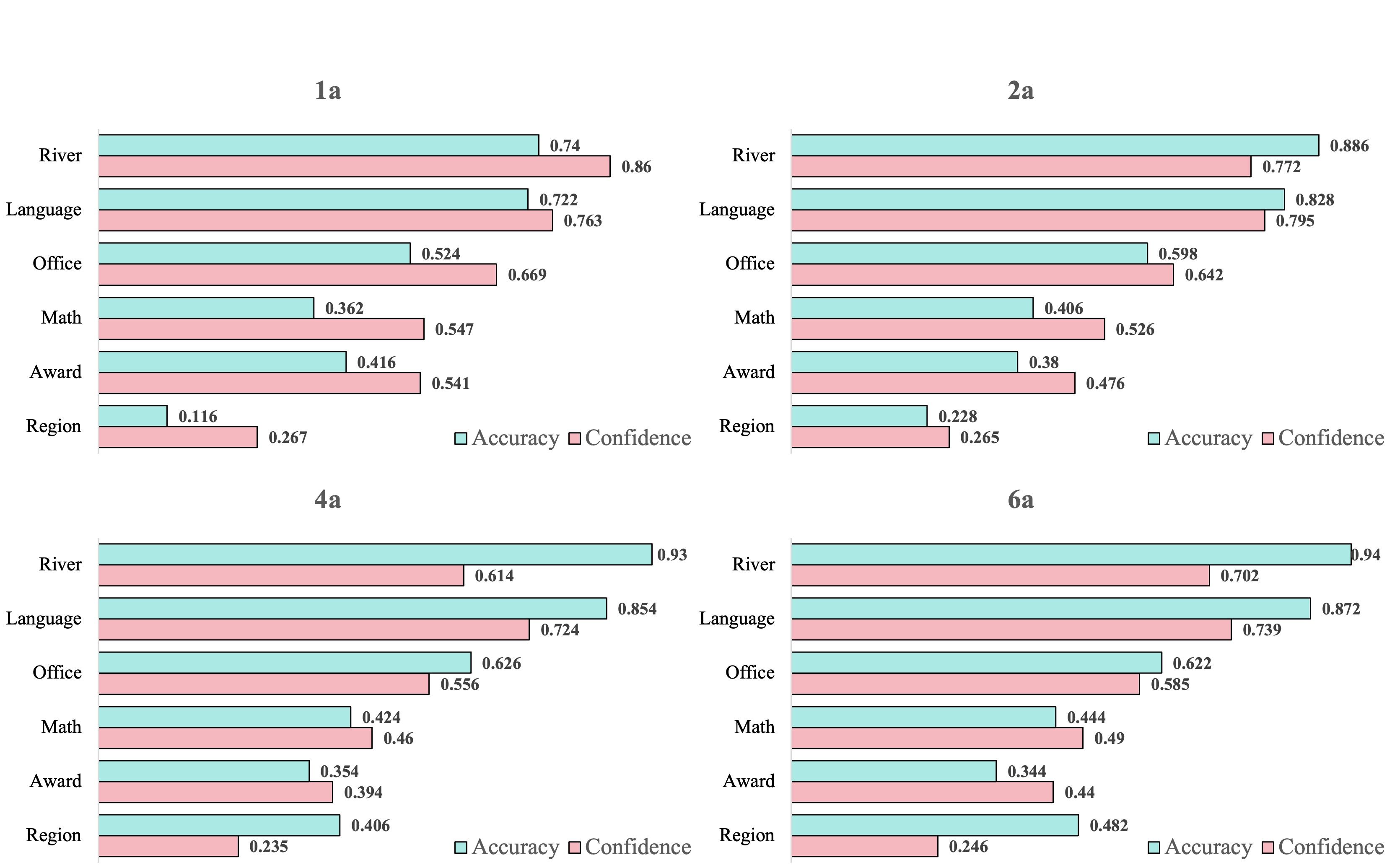}
    \caption{
    Domain-specific accuracy and confidence under different ground-truth sizes.
    }
    \label{fig:llama70B-avg-GT-Class-4gt}
\end{figure*}

\begin{figure}[t]
  \centering
  \includegraphics[width=\columnwidth]{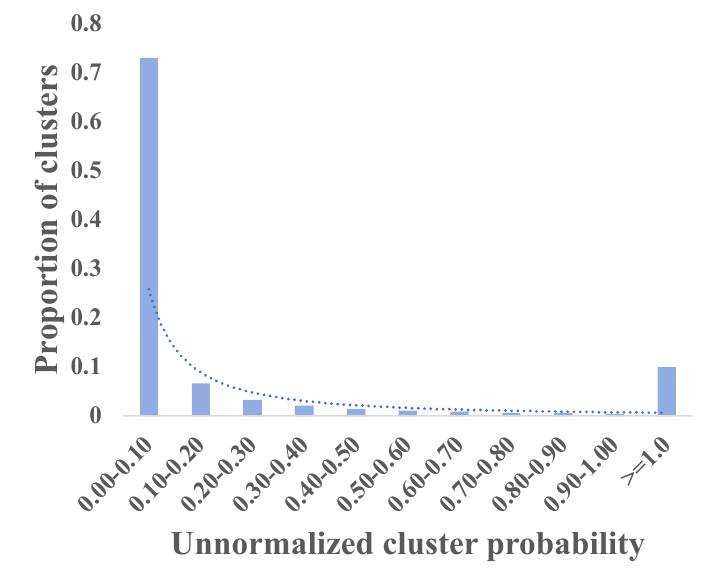}
  \caption{\textbf{Distribution of unnormalised cluster probabilities}
  $P_{\text{raw}}(C_m) = \sum_{r_j \in C_m} P(r_j)$ for
  \textbf{LLaMA-3.1-70B-Instruct}.
  Each bar shows the fraction of semantic clusters whose cumulative
  probability falls within the corresponding bin.
  Since probabilities are summed over $M\!=\!20$ samples without
  renormalisation, values can exceed~1.
  Over 70\% of clusters concentrate in $[0,\,0.1)$, motivating our
  filtering threshold $\tau\!=\!0.1$
  (see \S\ref{appendix:ClusterProbDist}).
  The dotted curve is a kernel density estimate for visual reference.}
  \label{fig:cluster-prob-dist}
\end{figure}

\subsection{Experiment Details}
\label{app:experiment-details}
\paragraph{Correctness Evaluation.}
Given the diversity of output formats, we use \texttt{GPT-4o} as our primary evaluator, following \cite{liu2023g}. An answer is deemed correct if GPT-4o determines that it matches any of the ground-truth answers. GPT-4o is also employed to assess semantic consistency between pairs of responses. To validate the reliability of these automatic judgments, we perform human evaluation on 300 randomly sampled instances, observing a Cohen’s~$\kappa$ of 0.92, which indicates high agreement among annotators.
\paragraph{Inference Framework.}
All experiments are conducted using
\textbf{vLLM}~\cite{kwon2023efficient} as the inference backend,
which provides efficient batched decoding and native access to
token-level log-probabilities required by probability-based confidence
estimators.

\paragraph{Sampling Configuration.}
We adopt two distinct sampling regimes depending on the confidence
estimation method:

\begin{itemize}[nosep,leftmargin=1.5em]
  \item \textbf{Single-sample methods}
        (verbalized confidence, post-hoc elicitation, self-verification,
         token probability, and accuracy evaluation):
        temperature is set to $T\!=\!0$ (greedy decoding) to obtain
        deterministic outputs.
  \item \textbf{Multi-sample methods}
        (consistency-based and semantic-entropy-based estimators):
        temperature is set to $T\!=\!1$ and we draw $M\!=\!20$ independent
        samples per question to capture the diversity of the model's
        output distribution.
\end{itemize}

\paragraph{Entropy-to-Confidence Conversion.}
For all entropy-based estimators (Prediction Entropy, Length-Normalised
Prediction Entropy, and Semantic Entropy), the raw entropy value $H$ is
converted into a confidence score via the exponential transformation
\[
  \mathrm{Conf} \;=\; e^{-H},
\]
as described in Eq.~\eqref{eq:entropy2conf}.
This ensures that all methods produce scores on a common $(0,\,1]$ scale,
enabling direct cross-method comparison.

\paragraph{LLM-based Evaluation.}
We use \texttt{GPT-4o} as the automated evaluator for two critical
components of our pipeline:

\begin{enumerate}[nosep,leftmargin=1.5em]
  \item \textbf{Answer Correctness Judgment.}
        Exact-match and rule-based heuristics perform poorly on our benchmark
        due to surface-form variations (e.g., ``U.S.A.'' vs.\ ``United States'',
        partial name matches, and aliasing across languages).
        We therefore prompt \texttt{GPT-4o} with the question, the model's
        response, and the ground-truth answer set, asking it to return a binary
        \texttt{True}/\texttt{False} verdict.

  \item \textbf{Semantic Similarity for Multi-sample Methods.}
        For consistency-based and semantic-entropy-based estimators,
        pairwise semantic equivalence between sampled responses must be
        determined to form semantic clusters.
        We again rely on \texttt{GPT-4o} to judge whether two responses
        convey the same factual answer, as embedding-based similarity alone
        cannot reliably distinguish near-miss paraphrases from genuinely
        distinct answers.
\end{enumerate}

\noindent
To validate the reliability of these automatic judgments, we perform human
evaluation on 300 randomly sampled instances (100 from answer correctness,
200 from pairwise similarity), observing a Cohen's~$\kappa$ of \textbf{0.92},
indicating near-perfect agreement between \texttt{GPT-4o} and human
annotators.

\paragraph{Aggregation Strategy.}
All reported metrics follow a \textbf{macro-averaging} scheme across the
six domains to prevent any single domain from dominating the aggregate.
The specific aggregation procedure differs by metric type:

\begin{enumerate}[nosep,leftmargin=1.5em]
  \item \textbf{Accuracy and Confidence.}
        For each answer-count setting ($k \in \{1,2,4,6\}$), we compute the
        metric independently within each domain and then report the
        unweighted mean across all six domains.
        For the \textsc{Math} domain, which is further subdivided into five
        number-type labels, we first average across labels before contributing
        a single domain-level score, ensuring that \textsc{Math} carries the
        same weight as any other domain.

  \item \textbf{AUROC.}
        Because AUROC requires a sufficient mixture of positive and negative
        instances, we first \emph{pool} all instances within a domain across
        the relevant answer-count settings---either a single $k$ when
        reporting per-setting AUROC, or all four settings
        ($k \in \{1,2,4,6\}$) when reporting the mixed-$k$ AUROC---and
        compute one AUROC value per domain on this pooled set.
        The final reported number is then the unweighted mean of the six
        domain-level AUROC scores.

  \item \textbf{Inter-domain averaging.}
        In both cases above, the final aggregate is an \emph{unweighted}
        mean over the six domain-level values.
        This macro-average treats every domain as an equally important
        evaluation scenario, preventing high-accuracy domains from inflating
        the overall metric
        (see \S\ref{app:domain-4gt} for further discussion).
\end{enumerate}


\subsection{Domain Effects: Relation Between Question Type and Difficulty}
\label{app:domain-4gt}

We analyze the domain-level calibration behavior of
\textbf{LLaMA-3.1-70B-Instruct} across varying ground-truth set sizes.
As illustrated in Figure~\ref{fig:llama70B-avg-GT-Class-4gt}, all
configurations (\texttt{1a/2a/4a/6a}) exhibit a highly consistent trend:
domains characterised by higher accuracy (lower intrinsic difficulty) also
demonstrate higher confidence levels.

\paragraph{Why Macro-Average as the Primary Metric.}
The six domains span a wide difficulty spectrum---from near-ceiling accuracy
in \textit{Language} and \textit{River} to substantially lower accuracy in
\textit{Region} and \textit{Award}.
If we were to report a \emph{micro-average} (i.e., pooling all instances
regardless of domain), the aggregate metric would be dominated by the
high-accuracy, low-difficulty domains, masking calibration deficiencies in
harder domains where reliable confidence estimation matters most.
To prevent this skew, we adopt the \textbf{macro-average}---computing each
metric independently within every domain and then averaging across
domains---as our primary reporting metric throughout the paper.
This treats every domain as an equally weighted evaluation scenario and
yields a more faithful picture of global calibration quality, especially
for safety-critical applications where a model's worst-case domain
performance is at least as important as its best-case performance.

\paragraph{Structural Origins of Difficulty Variation.}
Crucially, the difficulty gap across domains is \emph{not} an artefact of
inconsistent question phrasing---all six domains share the same
``Name one X that satisfies Y'' template
(Table~\ref{tab:core_prompts}; Appendix~\S\ref{appendix:QA-generation}).
Instead, it originates from the two fundamentally different
\emph{construction paths} used to populate multi-answer instances:

\begin{itemize}[nosep,leftmargin=1.5em]
  \item \textbf{Cross-subject Aggregation} (\textit{Language}, \textit{River}).
        Multiple subjects are composed into a single disjunctive query
        (e.g., ``Name one country where language~A, B, or~C is spoken'').
        Each disjunct maps to a small, well-separated set of countries,
        so the model only needs to recall \emph{any one} member from a union
        of compact, high-frequency fact sets.
        The resulting answer space is narrow and semantically unambiguous,
        yielding the \textbf{lowest difficulty}.

  \item \textbf{Intra-subject Expansion} (\textit{Award}, \textit{Office},
        \textit{Region}).
        A single subject's constraint window is widened (e.g., a broader year
        range for awards, a larger area interval for administrative divisions)
        to harvest $k$ ground-truth objects.
        Although each question still targets one coherent topic, the expanded
        window opens up a \emph{large candidate pool} of entities that share
        similar contextual attributes (e.g., overlapping tenures, comparable
        geographic areas).
        The model must perform fine-grained attribute linking---matching a
        specific temporal or spatial constraint to the correct entity among
        many plausible alternatives---which substantially increases
        \textbf{retrieval complexity and semantic confusability}.
\end{itemize}

\noindent
The \textit{\Math} domain occupies an intermediate position:
it bypasses factual retrieval entirely and instead requires recognition of
abstract numeric properties (primality, perfect squares, etc.) over
uniformly sampled integers.
This is cognitively simpler than the attribute-linking in Intra-subject
domains, yet harder than the disjunctive enumeration in Cross-subject
domains, because the model must apply rule-based reasoning rather than
surface-level association.

\medskip\noindent
In summary, the difficulty hierarchy 
is a direct consequence of the construction methodology described in
Appendix~\S\ref{appendix:QA-generation}.
Because difficulty is structurally determined by the domain's construction
path rather than by surface-level question variation, it remains stable
across all answer-count settings ($k \in \{1,2,4,6\}$).

\section{Results and Analysis}
\subsection{Consistent Trends Across Models and Scales.}
\label{appendix:all model result}

We report the \textbf{confidence} of all evaluated models under different numbers of valid answers (\texttt{1a/2a/4a/6a}), as shown in Tables~\ref{tab:confidence_qwen257b}--\ref{tab:confidence_deepseekv3}. Note that for API-based models (e.g., GPT-4o, GPT-4o-mini, and DeepSeek-V3), we do not report methods that rely on logit probabilities or internal token distributions, such as Prob-Entropy, N-Prob-Entropy and Sem Entropy. For entropy-based methods (Prob Entropy, N-Prob Entropy, and Sem Entropy), we convert entropy $H$ into a confidence score via $\text{conf} = e^{-H}$ to ensure a consistent and comparable scale with other methods, where higher values uniformly indicate greater confidence.

Across model families and parameter scales, a clear and consistent pattern emerges: model accuracy follows a steady upward trend as the number of valid answers increases. However, the model confidence exhibits a diametrically opposite decline over the same range. This diverging behavior reveals a significant misalignment between the models' actual performance and their certainty: as the task environment becomes more "favorable" (i.e., more valid answers are available), the models paradoxically become increasingly "underconfident." Such a monotonic decline in confidence despite rising accuracy demonstrates that existing calibration methods are unable to remain stable when confronted with multi-answer variation, highlighting a fundamental limitation of current approaches.

\subsection{Why Does Confidence Decline? A Per-Method Analysis}
\label{app:method-analysis}

The consistent confidence decline observed across all method families
(§\ref{appendix:all model result}) naturally raises the question: \emph{what
is the underlying mechanism for each category of methods?}
We analyze the three major families separately below.

\paragraph{Consistency-based methods.}
Methods such as Consistency and Semantic Entropy estimate confidence by
measuring agreement among multiple sampled responses. Through case-level
inspection, we find that the confidence drop is directly attributable to
\textbf{answer diversity fragmenting the generation probability mass}.
When a question has a single valid answer, the model's sampled responses
tend to converge on one surface form, yielding high inter-sample agreement.
As the number of valid answers grows, the model distributes its
generations across multiple equally correct but textually distinct answers.
For instance, when asked \emph{``Name one person who received the Nobel
Prize in Physics in 1995,''} the model alternates among ``Martin Lewis
Perl'' and ``Frederick Reines''---both correct---across samples.
Consistency-based methods interpret this diversity as disagreement,
producing a low confidence score that is indistinguishable from genuine
uncertainty caused by alternation between correct and incorrect answers.

\paragraph{Probability-based methods.}
For generation-probability methods (e.g., Perplexity), the confidence
decline might appear surprising at first, since these methods operate on a
single greedy or sampled response rather than on multi-sample agreement.
To investigate the root cause, we measure the entropy of the \textbf{first
generated token's top-20 logits} for Perplexity on LLaMA-3.1-70B-Instruct,
stratified by answer count:

\begin{table}[h]
\centering
\small
\begin{tabular}{lcccc}
\toprule
& \textbf{1a} & \textbf{2a} & \textbf{4a} & \textbf{6a} \\
\midrule
First-token entropy & 1.22 & 1.30 & 1.53 & 1.68 \\
\bottomrule
\end{tabular}
\caption{Average entropy over the top-20 logits of the first generated
token under different answer counts (LLaMA-3.1-70B-Instruct).}
\label{tab:first-token-entropy}
\end{table}

The monotonic increase in first-token entropy confirms that, even at the
very onset of generation, the model's probability mass is more dispersed
when more valid answers exist. Intuitively, the model ``hesitates'' among
multiple plausible starting tokens (e.g., ``Martin'' vs.\ ``Frederick''),
each leading to a different correct answer. This initial hesitation
propagates through the entire sequence, resulting in lower overall
sequence probability and thus lower Perplexity-based confidence.
In other words, the same answer-diversity mechanism that afflicts
consistency-based methods also affects probability-based methods, but
manifests at the \emph{token distribution} level rather than the
\emph{sample agreement} level.

\paragraph{Double-turn methods.}
In contrast to the above two families, double-turn methods
(e.g., P\textsubscript{True}, Self-Ask) exhibit a noticeably smaller
confidence decline across answer-count settings.
As shown in Table~\ref{tab:confidence_llama70b}, P\textsubscript{True}-Consis-Cand
maintains confidence scores of 88.1, 87.9, 87.1, and 87.9 for 1a through
6a, respectively---a variation of less than 1.2 points.
This robustness stems from the fact that double-turn methods
\textbf{decouple confidence estimation from the generation process}.
Rather than deriving confidence from how the answer was produced
(token probabilities or sampling agreement), these methods pose a
separate verification query---e.g., \emph{``Is the following answer
correct: Martin Lewis Perl? (True/False)''}---and assess confidence based
on the model's \emph{judgment} ability.
Since verifying a single correct answer does not require the model to
choose among alternatives, the verification step is largely unaffected
by how many other valid answers exist.
However, this advantage comes at the cost of an additional inference pass
per question, and as shown in our main results
(Tables~\ref{tab:multi-answer-auroc-llama70b}--\ref{tab:multi-answer-auroc-llama8b}),
double-turn methods do not consistently outperform our proposed SCA in
AUROC, which achieves competitive or superior calibration while remaining
a single-turn approach.

\begin{figure}[t]
    \centering
    \includegraphics[width=\linewidth]{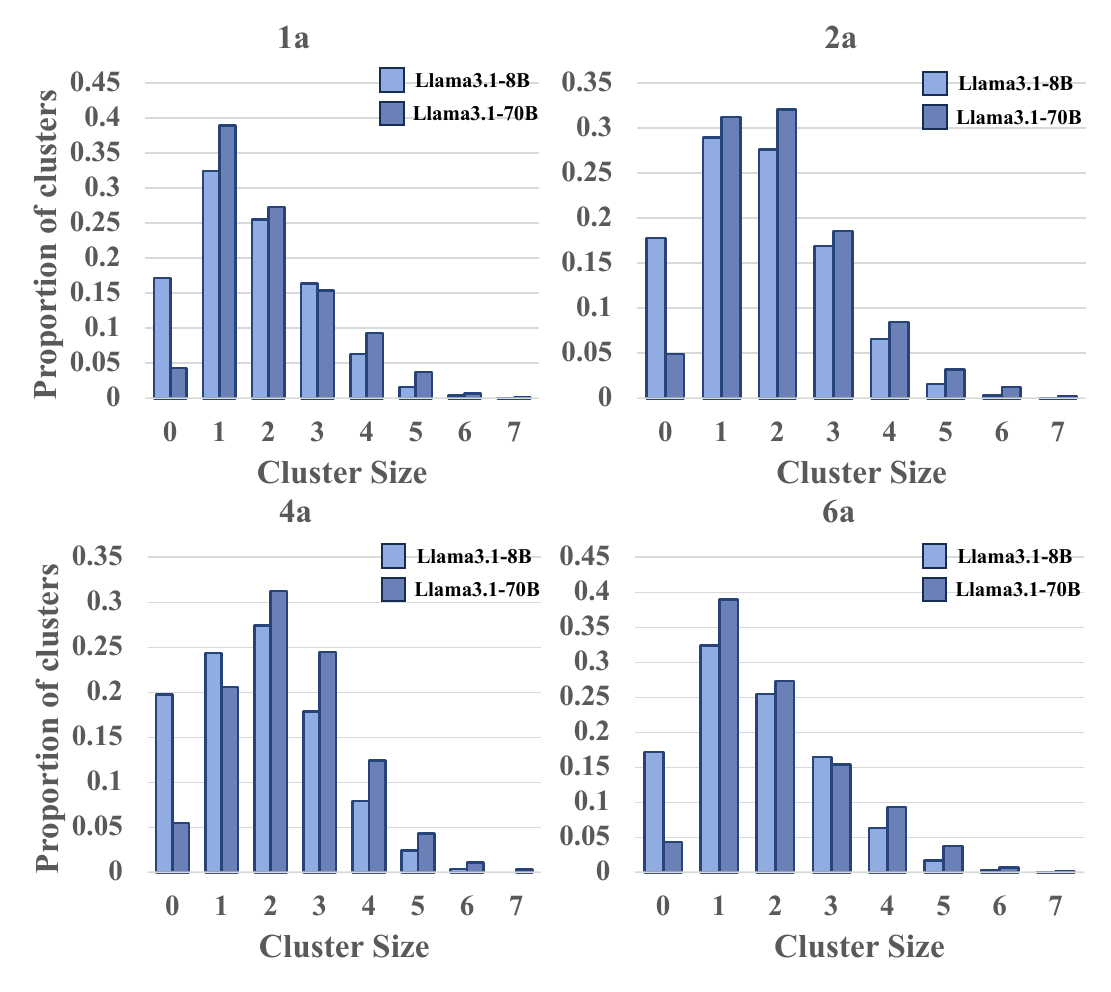}
    \caption{
\textbf{Distribution of  cluster sizes of {llama} family.}
}
    \label{fig:cluster_num_distribution}
    \vspace{-0.5cm}
\end{figure}

\subsection{Analysis of Cluster Probability Distribution}

\label{appendix:ClusterProbDist}
We apply a token-level probability-based filter and retain only clusters whose cumulative probability exceeds a threshold of $\tau=0.1$.
\paragraph{Why Unnormalised Cluster Probabilities.}
A key design decision in our semantic-entropy pipeline is to measure each
cluster's importance by its \textbf{unnormalised cumulative probability},
defined as the sum of token-level generation probabilities across all
member responses:
\[
  P_{\text{raw}}(C_m) \;=\; \sum_{r_j \in C_m} P(r_j),
\]
where $P(r_j)$ is the sequence-level probability of the $j$-th sampled
response.
Because we draw $M\!=\!20$ samples per question, the raw sum
$P_{\text{raw}}(C_m)$ is \emph{not} constrained to $[0,1]$---clusters that
capture the model's dominant output mode can accumulate values exceeding~1.

We deliberately avoid renormalising these probabilities across clusters
(i.e., converting them into a simplex distribution $\hat{P}(C_m) =
P_{\text{raw}}(C_m) / \sum_m P_{\text{raw}}(C_m)$).
The reason is that normalisation would \emph{inflate} the relative weight
of low-probability clusters: even when \emph{all} clusters carry negligible
absolute mass (e.g., the model is highly uncertain and scatters probability
across many fragmented outputs), normalisation would still assign some
cluster a large \emph{relative} share, creating the false impression of a
confident semantic mode.
By operating on unnormalised probabilities, we preserve the absolute scale
of the model's output distribution, so that a cluster's importance directly
reflects the model's genuine probability commitment to that semantic
content.

\paragraph{Empirical Distribution and Threshold Selection.}
Figure~\ref{fig:cluster-prob-dist} shows the distribution of
$P_{\text{raw}}(C_m)$ across all clusters produced by
\textbf{LLaMA-3.1-70B-Instruct}.
The distribution exhibits a striking \textbf{bimodal} structure:

\begin{itemize}[nosep,leftmargin=1.5em]
  \item A dominant spike in the $[0,\,0.1)$ bin, accounting for over
        \textbf{70\%} of all clusters.
        These are low-mass clusters that each capture only a tiny fraction
        of the 20 samples---typically one or two responses with individually
        low generation probabilities.
        Manual inspection confirms that they predominantly consist of
        fragmented, degenerate, or semantically irrelevant outputs.
  \item A secondary concentration at $P_{\text{raw}} \!\ge\! 1.0$
        (approximately \textbf{10\%} of clusters), representing the model's
        dominant answer modes where multiple high-probability samples converge
        on the same semantic content.
\end{itemize}

\noindent
Between these two modes, the bins from $[0.1,\,1.0)$ are sparsely
populated, forming a clear \textbf{probability gap} that provides a natural
decision boundary.
We therefore set the filtering threshold at
\[
  \tau \;=\; 0.1,
\]
retaining only clusters with $P_{\text{raw}}(C_m) \ge 0.1$.
The rationale for this specific value is threefold:

\begin{enumerate}[nosep,leftmargin=1.5em]
  \item \textbf{Empirical gap alignment.}
        The threshold sits precisely at the boundary between the dominant
        near-zero spike and the sparsely populated mid-range bins.
        Any value in the interval $[0.05,\,0.15]$ would produce similar
        filtering outcomes; $0.1$ is chosen as the round value at the centre
        of this stable region.
  \item \textbf{Intuitive interpretation under 20 samples.}
        With $M\!=\!20$ draws, $P_{\text{raw}}(C_m) = 0.1$ corresponds
        roughly to a cluster that either (a) contains a single sample with
        generation probability $\approx 0.1$, or (b) aggregates two samples
        each with probability $\approx 0.05$.
        In both cases the cluster represents less than $0.5\%$ of the total
        probability budget ($\sum_m P_{\text{raw}}(C_m)$), making it
        negligible for uncertainty estimation.
  \item \textbf{Cross-domain and cross-model stability.}
        We verified that the $[0,\,0.1)$ spike persists across all six domains
        and all evaluated models, confirming that $\tau\!=\!0.1$ is not
        over-fitted to a single experimental condition.
\end{enumerate}

\noindent
Applying this threshold yields a compact set of \emph{effective} semantic
clusters that faithfully represent the model's meaningful output modes,
filtering out sampling noise and providing a cleaner basis for computing
semantic entropy and analysing confidence--rank dynamics.



\begin{table}[t]
\centering
\small
\caption{\textbf{Spearman correlation of confidence with {\answerspace}}  across ground-truth(GT) set sizes using \textit{consistency} method.} 
\label{tab:spearman_results_v2_conf_only}
\begin{tabular}{lcccc}
\toprule
Model & \textbf{1a} & \textbf{2a} & \textbf{4a} & \textbf{6a} \\
\midrule
Qwen2.5-7B    & 0.009  & -0.000 & -0.000 & -0.012 \\
Qwen2.5-14B   & -0.443 & -0.438 & -0.427 & -0.421 \\
Qwen2.5-32B   & -0.421 & -0.421 & -0.407 & -0.399 \\
Qwen2.5-72B   & -0.616 & -0.551 & -0.533 & -0.516 \\
Llama-3.1-8B  & 0.242  & 0.227  & 0.325  & 0.366  \\
Llama-3.1-70B & -0.393 & -0.314 & -0.167 & -0.055 \\
GPT-4o-mini   & -0.590 & -0.563 & -0.567 & -0.544 \\
GPT-4o        & -0.821 & -0.779 & -0.729 & -0.684 \\
DeepSeek-V3   & -0.854 & -0.827 & -0.820 & -0.807 \\
\bottomrule
\end{tabular}
\end{table}

\subsection{Spearman metric cross Knowledge Coverage}
\label{app:Spearman Validity}

\paragraph{Validity of Spearman Correlation.}
Spearman's rank correlation is sensitive to tied ranks and can produce
misleading results when the distribution of one variable is heavily
concentrated at a single value.
We therefore verify that the effective answer-space size (i.e., the number
of semantic clusters retained after filtering) is sufficiently dispersed
to support a meaningful rank-based analysis.

Figure~\ref{fig:cluster_num_distribution} shows the distribution of cluster
sizes across both models and all four ground-truth settings
(\texttt{1a/2a/4a/6a}) after applying the probability threshold
$\tau\!=\!0.1$.
Two key observations confirm the validity of Spearman correlation
in our setting:

\begin{enumerate}[nosep,leftmargin=1.5em]
  \item \textbf{No single cluster size dominates.}
        In every setting, cluster sizes span a range from 0 to 6+, and no
        individual bin accounts for more than 40\% of the total mass.
        This rules out the degenerate case where most instances collapse
        to the same cluster count---which would produce excessive tied ranks
        and inflate or deflate the correlation coefficient.

  \item \textbf{Systematic shift with increasing ground-truth cardinality.}
        As the number of ground-truth answers grows from $k\!=\!1$ to
        $k\!=\!6$, the cluster-size distribution shifts smoothly toward
        larger values without collapsing into a narrow mode or generating
        extreme outliers.
        This gradual expansion confirms that the effective answer space
        grows in a stable manner, providing a well-graded ordinal variable
        that Spearman correlation can reliably capture.
\end{enumerate}

\paragraph{Correlation Results.}
Table~\ref{tab:spearman_results_v2_conf_only} reports the Spearman
correlation between model confidence and knowledge coverage using \textit{consistency} method.
Across multiple models, we observe a consistent \textbf{negative
correlation}: as the effective answer space expands (i.e., more
semantically distinct clusters survive filtering), the model's reported
confidence \emph{decreases}.
This finding indicates that higher confidence does not reflect broader
factual coverage; rather, models tend to be most confident when their
output distribution concentrates on a small number of semantic modes,
even though a wider set of valid answers exists.

Notably, \textbf{larger models exhibit a stronger negative correlation}
than their smaller counterparts.
This suggests that as model capacity increases, the confidence score
becomes \emph{more} sensitive to the expansion of the effective answer
space---precisely the regime where miscalibration is most consequential.
This observation is consistent with our earlier findings
(Section~\ref{sec:results}), where larger models display steeper
confidence degradation as the number of ground-truth answers grows,
reinforcing the conclusion that scale alone does not mitigate---and may
even exacerbate---the confidence--coverage misalignment under
multi-answer settings.

\section{Semantic Confidence Aggregation}
\label{app:MAPS}

\subsection{Generalization Across Models}
\label{app:SCA auroc generalization}

To evaluate whether our method generalises beyond the primary experimental
models, we extend the full evaluation suite to additional architectures
and scales:
\textbf{Qwen-2.5-7B-Instruct} (Table~\ref{tab:multi-answer-auroc-qwen7b}),
\textbf{Qwen-2.5-14B-Instruct} (Table~\ref{tab:multi-answer-auroc-qwen14b}),
\textbf{Qwen-2.5-32B-Instruct} (Table~\ref{tab:multi-answer-auroc-qwen32b}),
\textbf{Qwen-2.5-72B-Instruct} (Table~\ref{tab:multi-answer-auroc-qwen}),
and \textbf{LLaMA-3.1-8B-Instruct} (Table~\ref{tab:multi-answer-auroc-llama8b}).

Across the majority of models and settings, SCA achieves the best or
second-best AUROC.
On the smaller-scale models (7B and 14B) and LLaMA-8B, SCA dominates
all baselines under every answer-mixture configuration, and its advantage
becomes more pronounced as the answer mixture broadens
(i.e., moving from \textbf{[1]} to \textbf{[1,2,4,6]}), confirming that
the method's strength lies precisely in the multi-answer regime where
existing estimators degrade most severely.


On the 72B Qwen and 70B LLaMA models, SCA again achieves the best or
near-best AUROC in most settings, demonstrating that its effectiveness
is not limited to smaller-scale models.

Furthermore, SCA with $\tau\!=\!0$ (no filtering) performs on par with or
better than the threshold-tuned variant in nearly all cases across all
models, suggesting that the method is robust to the choice of filtering
threshold and does not require per-model hyper-parameter tuning.

\paragraph{Statistical Significance.}
To rigorously assess whether SCA's improvements are statistically
meaningful, we conduct a bootstrap significance test.
For each answer-mixture setting, we randomly sample 1{,}000 instances
with replacement from the corresponding test pool and compute the AUROC
for both SCA and every baseline method.
This resampling procedure is repeated 1{,}000 times, yielding an empirical
distribution of AUROC differences.
A baseline is marked with $\dagger$ in
Tables~\ref{tab:multi-answer-auroc-qwen7b}--\ref{tab:multi-answer-auroc-llama8b}
if the two-sided $p$-value exceeds 0.05 (i.e., the difference from SCA is
\emph{not} statistically significant).
Across all models and settings, the vast majority of baselines show
$p < 0.05$, confirming that SCA's advantage is statistically significant
in the overwhelming majority of comparisons.
The few non-significant cases (marked $\dagger$) occur mostly in the
single-answer setting \textbf{[1]}, where the multi-answer challenge is
absent and several strong baselines already perform near-optimally.

These results, combined with the primary findings on
LLaMA-3.1-70B-Instruct (Table~\ref{tab:multi-answer-auroc-llama70b}),
demonstrate that SCA is \textbf{model-agnostic} and
\textbf{scale-agnostic}, maintaining strong performance across different
model families (LLaMA, Qwen) and parameter scales (7B--72B).
While SCA is the clear overall winner as a \emph{single-turn} calibration
method, the 32B results highlight that sufficiently capable models may
benefit from double-turn probing when the additional inference cost is
acceptable---an interesting direction for future work on adaptive
confidence estimation.

\subsection{Results on NQ Dataset}
\label{app:nq_results}

To further assess the generalization of SCA on single-answer datasets, we randomly sample 500 questions from NQ-open~\cite{kwiatkowski2019natural} and evaluate calibration performance (AUROC). The results are shown in Table~\ref{tab:nq-results}.

\begin{table}[t]
\centering
\small
\setlength{\tabcolsep}{4pt}
\renewcommand{\arraystretch}{0.95}
\caption{
\textbf{AUROC score on NQ-open} (500 randomly sampled questions) of different calibration methods on LLaMA-3.1-70B-Instruct.
$\tau=0$ denotes no-filtering, while $\tau\neq0$ uses the threshold value tuned on the development set.
\textbf{Bold} marks the best, \underline{underlined} the second-best.
}
\begin{tabular}{lc}
\toprule
\textbf{Method} & \textbf{NQ} \\
\midrule
\multicolumn{2}{l}{\textbf{Single-turn}} \\
\midrule
\textit{Question-level} & \\
Prob Entropy            & 75.9 \\
N-Prob Entropy          & \underline{76.2} \\
Sem Entropy             & 76.0 \\
\midrule
\textit{Answer-level} & \\
Verb                    & 64.0 \\
Verb-Topk               & 69.5 \\
Consis                  & \textbf{78.0} \\
Consis-Verb             & 73.3 \\
Consis-Verb-Topk        & 74.6 \\
Perplexity              & 75.7 \\
\midrule
\multicolumn{2}{l}{\textbf{Double-turn}} \\
\midrule
P\textsubscript{True}-Consis        & 65.1 \\
P\textsubscript{True}-Prob          & 70.1 \\
P\textsubscript{True}-Consis-Cand   & 59.0 \\
P\textsubscript{True}-Prob-Cand     & 72.4 \\
Self-Ask                            & 67.4 \\
Self-Ask-Cand                       & 67.8 \\
\midrule
\multicolumn{2}{l}{\textbf{Confidence Aggregation Baselines}} \\
\midrule
{SNCA ($\tau=0.15$)}    & 74.8 \\
{SNCA ($\tau=0$)}       & 50.0 \\
{SFCA ($\tau=0.35$)}    & 73.7 \\
{SFCA ($\tau=0$)}       & 50.0 \\
\midrule
\multicolumn{2}{l}{\textbf{Ours}} \\
\midrule
{SCA ($\tau=0.9$)}      & \underline{76.2} \\
{SCA ($\tau=0$)}        & 75.6 \\
\bottomrule
\end{tabular}
\label{tab:nq-results}
\end{table}

SCA ($\tau=0$) achieves 75.6 AUROC without any threshold tuning, outperforming all double-turn methods and confidence aggregation baselines. With an optimal threshold ($\tau=0.9$), SCA matches the best single-turn question-level method (N-Prob Entropy, 76.2), demonstrating strong generalization on single-answer benchmarks.


\begin{table}[t]
\centering
\small
\setlength{\tabcolsep}{4pt}
\renewcommand{\arraystretch}{0.95}
\caption{
\textbf{AUROC score on {Qwen-2.5-7B-Instruct}} of different calibration methods under increasing answer mixture.
$\tau=0$ denotes no-filtering, while $\tau\neq0$ uses the threshold value  tuned on the development set.
\textbf{Bold} marks the best, \underline{underlined} the second-best in each column.
$\dagger$ indicates the difference from our SCA method is not statistically significant ($p > 0.05$).
}
\begin{tabular}{lcccc}
\toprule
\textbf{Method} & \textbf{[1]} & \textbf{[1,2]} & \textbf{[1,2,4]} & \textbf{[1,2,4,6]} \\
\midrule
\multicolumn{5}{l}{\textbf{Single-turn}} \\
\midrule
\textit{Question-level} & & & & \\
Prob Entropy                                    & 77.7 & 77.5 & 75.7 & 74.4 \\
N-Prob Entropy                                  & 77.5$^\dagger$ & 76.1 & 74.3 & 73.6 \\
Sem Entropy                                     & 75.6 & 75.3 & 73.4 & 72.1 \\
\midrule
\textit{Answer-level} & & & & \\
Verb                                            & 58.3 & 57.0 & 56.4 & 55.2 \\
Verb-Topk                                       & 61.7 & 56.9 & 52.8 & 51.8 \\
Consis                                          & 78.4 & 78.0 & 76.3 & 75.3 \\
Consis-Verb                                     & 75.5 & 74.7 & 74.3 & 73.7 \\
Consis-Verb-Topk                                & 75.6 & 74.2 & 73.6 & 72.4 \\
Perplexity                                      & 78.8$^\dagger$ & 77.9 & 76.0 & 75.4 \\

\midrule
\multicolumn{5}{l}{\textbf{Double-turn}} \\
\midrule
P\textsubscript{True}-Consis                    & 71.3 & 67.3 & 67.8 & 67.3 \\
P\textsubscript{True}-Prob                      & 76.2$^\dagger$ & 71.7 & 72.2 & 71.0 \\
P\textsubscript{True}-Consis-Cand               & 69.0 & 69.3 & 68.0 & 67.0 \\
P\textsubscript{True}-Prob-Cand                 & 77.6$^\dagger$ & 76.6 & 74.6 & 73.3 \\
Self-Ask                                        & 69.2 & 65.9 & 64.9 & 62.8 \\
Self-Ask-Cand                                   & 76.5 & 75.0 & 72.7 & 71.3 \\
\midrule
\multicolumn{5}{l}{\textbf{Confidence Aggregation Baselines}} \\
\midrule
{SNCA ($\tau=0.2$)}                              & 74.1 & 74.5 & 72.7 & 71.6 \\
{SNCA ($\tau=0$)}                                & 50.0 & 50.0 & 50.0 & 50.0 \\
{SFCA ($\tau=0.15$)}                             & 76.4 & 76.0 & 75.5 & 74.3 \\
{SFCA ($\tau=0$)}                                & 50.0 & 50.0 & 50.0 & 50.0 \\
\midrule
\multicolumn{5}{l}{\textbf{Ours}} \\
\midrule
{SCA ($\tau=0.1$)}                              & \underline{79.9} & \underline{79.7} & \underline{78.1} & \underline{77.0} \\
{SCA ($\tau=0$)}                                & \textbf{80.2} & \textbf{79.8} & \textbf{78.2} & \textbf{77.1} \\

\bottomrule
\end{tabular}
\label{tab:multi-answer-auroc-qwen7b}
\end{table}

\begin{table}[t]
\centering
\small
\setlength{\tabcolsep}{4pt}
\renewcommand{\arraystretch}{0.95}
\caption{
\textbf{AUROC score on {Qwen-2.5-14B-Instruct}} of different calibration methods under increasing answer mixture.
$\tau=0$ denotes no-filtering, while $\tau\neq0$ uses the threshold value  tuned on the development set.
\textbf{Bold} marks the best, \underline{underlined} the second-best in each column.
$\dagger$ indicates the difference from our SCA method is not statistically significant ($p > 0.05$).
}
\begin{tabular}{lcccc}
\toprule
\textbf{Method} & \textbf{[1]} & \textbf{[1,2]} & \textbf{[1,2,4]} & \textbf{[1,2,4,6]} \\
\midrule
\multicolumn{5}{l}{\textbf{Single-turn}} \\
\midrule
\textit{Question-level} & & & & \\
Prob Entropy                                    & 81.0 & 78.6 & 76.5 & 75.3 \\
N-Prob Entropy                                  & 81.3 & 79.0 & 77.0 & 75.3 \\
Sem Entropy                                     & 77.8 & 77.6 & 75.6 & 74.2 \\
\midrule
\textit{Answer-level} & & & & \\
Verb                                            & 69.1 & 65.6 & 62.3 & 60.1 \\
Verb-Topk                                       & 60.6 & 57.0 & 53.5 & 51.5 \\
Consis                                          & 78.6 & 78.0 & 76.4 & 75.2 \\
Consis-Verb                                     & 74.1 & 72.6 & 71.4 & 69.9 \\
Consis-Verb-Topk                                & 78.1 & 76.5 & 74.6 & 73.5 \\
Perplexity                                      & 81.6 & 79.8 & \underline{77.6} & \underline{76.0} \\

\midrule
\multicolumn{5}{l}{\textbf{Double-turn}} \\
\midrule
P\textsubscript{True}-Consis                    & 74.3 & 72.8 & 72.1 & 71.1 \\
P\textsubscript{True}-Prob                      & 78.2 & 76.6 & 76.2 & 75.1 \\
P\textsubscript{True}-Consis-Cand               & 69.6 & 67.8 & 67.3 & 66.8 \\
P\textsubscript{True}-Prob-Cand                 & 79.4 & 77.0 & 75.3 & 74.2 \\
Self-Ask                                        & 67.0 & 65.7 & 65.4 & 64.8 \\
Self-Ask-Cand                                   & 74.9 & 72.9 & 70.2 & 69.1 \\
\midrule
\multicolumn{5}{l}{\textbf{Confidence Aggregation Baselines}} \\
\midrule
{SNCA ($\tau=0.4$)}                              & 76.7 & 76.6 & 74.7 & 73.3 \\
{SNCA ($\tau=0$)}                                & 50.0 & 50.0 & 50.0 & 50.0 \\
{SFCA ($\tau=0.3$)}                              & 78.5 & 77.7 & 75.6 & 74.1 \\
{SFCA ($\tau=0$)}                                & 50.0 & 50.0 & 50.0 & 50.0 \\
\midrule
\multicolumn{5}{l}{\textbf{Ours}} \\
\midrule
{SCA ($\tau=0.1$)}                              & \textbf{82.7} & \textbf{81.0} & \textbf{78.7} & \textbf{77.1} \\
{SCA ($\tau=0$)}                                & \underline{82.6} & \underline{80.9} & \textbf{78.7} & \textbf{77.1} \\

\bottomrule
\end{tabular}
\label{tab:multi-answer-auroc-qwen14b}
\end{table}

\begin{table}[t]
\centering
\small
\setlength{\tabcolsep}{4pt}
\renewcommand{\arraystretch}{0.95}
\caption{
\textbf{AUROC score on {Qwen-2.5-32B-Instruct}} of different calibration methods under increasing answer mixture.
$\tau=0$ denotes no-filtering, while $\tau\neq0$ uses the threshold value  tuned on the development set.
\textbf{Bold} marks the best, \underline{underlined} the second-best in each column.
$\dagger$ indicates the difference from our SCA method is not statistically significant ($p > 0.05$).
}
\begin{tabular}{lcccc}
\toprule
\textbf{Method} & \textbf{[1]} & \textbf{[1,2]} & \textbf{[1,2,4]} & \textbf{[1,2,4,6]} \\
\midrule
\multicolumn{5}{l}{\textbf{Single-turn}} \\
\midrule
\textit{Question-level} & & & & \\
Prob Entropy                                    & 77.9 & 74.2 & 73.5 & 72.0 \\
N-Prob Entropy                                  & 78.3 & 74.6 & 73.5 & 72.0 \\
Sem Entropy                                     & 73.9 & 70.5 & 70.5 & 69.5 \\
\midrule
\textit{Answer-level} & & & & \\
Verb                                            & 62.8 & 60.3 & 59.4 & 58.7 \\
Verb-Topk                                       & 58.9 & 55.2 & 51.6 & 49.3 \\
Consis                                          & 75.3 & 72.3 & 72.4 & 71.3 \\
Consis-Verb                                     & 73.7 & 70.7 & 69.8 & 68.7 \\
Consis-Verb-Topk                                & 75.5 & 73.7 & 72.8 & 71.4 \\
Perplexity                                      & \textbf{79.2} & 75.8 & 74.7 & 73.1 \\

\midrule
\multicolumn{5}{l}{\textbf{Double-turn}} \\
\midrule
P\textsubscript{True}-Consis                    & 69.5 & 70.5 & 69.8 & 69.7 \\
P\textsubscript{True}-Prob                      & 79.0 & \textbf{78.2} & \textbf{77.3} & \textbf{77.0} \\
P\textsubscript{True}-Consis-Cand               & 66.8 & 67.0 & 66.4 & 66.4 \\
P\textsubscript{True}-Prob-Cand                 & 78.3 & \underline{76.4} & 75.0 & \underline{74.8} \\
Self-Ask                                        & 72.4 & 70.5 & 68.3 & 67.1 \\
Self-Ask-Cand                                   & 72.6 & 69.8 & 67.5 & 66.4 \\
\midrule
\multicolumn{5}{l}{\textbf{Confidence Aggregation Baselines}} \\
\midrule
{SNCA ($\tau=0.65$)}                             & 72.9 & 69.1 & 69.3 & 68.3 \\
{SNCA ($\tau=0$)}                                & 50.0 & 50.0 & 50.0 & 50.0 \\
{SFCA ($\tau=0.35$)}                             & 73.8 & 71.9 & 71.5 & 70.3 \\
{SFCA ($\tau=0$)}                                & 50.0 & 50.0 & 50.0 & 50.0 \\
\midrule
\multicolumn{5}{l}{\textbf{Ours}} \\
\midrule
{SCA ($\tau=0.05$)}                             & \textbf{79.2} & 76.1 & \underline{75.3} & 73.8 \\
{SCA ($\tau=0$)}                                & \underline{79.1} & 76.1 & \underline{75.3} & 73.8 \\

\bottomrule
\end{tabular}
\label{tab:multi-answer-auroc-qwen32b}
\end{table}

\begin{table}[t]
\centering
\small
\setlength{\tabcolsep}{4pt}
\renewcommand{\arraystretch}{0.95}
\caption{
\textbf{AUROC score on {Qwen-2.5-72B-Instruct}} of different calibration methods under increasing answer mixture.
$\tau=0$ denotes no-filtering, while $\tau\neq0$ uses the threshold value  tuned on the development set.
\textbf{Bold} marks the best, \underline{underlined} the second-best in each column.
$\dagger$ indicates the difference from our SCA method is not statistically significant ($p > 0.05$).
}
\begin{tabular}{lcccc}
\toprule
\textbf{Method} & \textbf{[1]} & \textbf{[1,2]} & \textbf{[1,2,4]} & \textbf{[1,2,4,6]} \\
\midrule
\multicolumn{5}{l}{\textbf{Single-turn}} \\
\midrule
\textit{Question-level} & & & & \\
Prob Entropy                                    & \textbf{82.4}$^\dagger$ & 79.1 & 78.4 & \underline{77.5} \\
N-Prob Entropy                                  & 81.8$^\dagger$ & 78.4 & 77.0 & 76.2 \\
Sem Entropy                                     & 77.3 & 75.6 & 75.1 & 73.9 \\
\midrule
\textit{Answer-level} & & & & \\
Verb                                            & 64.2 & 60.6 & 58.5 & 57.4 \\
Verb-Topk                                       & 67.7 & 64.1 & 59.9 & 57.7 \\
Consis                                          & 78.2 & 77.1 & 76.7 & 75.8 \\
Consis-Verb                                     & 77.5 & 75.8 & 75.4 & 74.1 \\
Consis-Verb-Topk                                & 75.1 & 73.2 & 72.2 & 71.0 \\
Perplexity                                      & \underline{82.0}$^\dagger$ & 79.9$^\dagger$ & \underline{78.5} & 77.4 \\

\midrule
\multicolumn{5}{l}{\textbf{Double-turn}} \\
\midrule
P\textsubscript{True}-Consis                    & 73.7 & 70.0 & 69.4 & 69.6 \\
P\textsubscript{True}-Prob                      & 81.3$^\dagger$ & 77.2 & 76.2 & 76.3 \\
P\textsubscript{True}-Consis-Cand               & 71.1 & 69.2 & 68.3 & 68.2 \\
P\textsubscript{True}-Prob-Cand                 & 81.5$^\dagger$ & 77.5 & 75.8 & 75.6 \\
Self-Ask                                        & 67.1 & 64.1 & 62.1 & 61.6 \\
Self-Ask-Cand                                   & 74.4 & 70.5 & 67.2 & 66.0 \\
\midrule
\multicolumn{5}{l}{\textbf{Confidence Aggregation Baselines}} \\
\midrule
{SNCA ($\tau=0.2$)}                              & 76.5 & 74.9 & 73.5 & 72.0 \\
{SNCA ($\tau=0$)}                                & 50.0 & 50.0 & 50.0 & 50.0 \\
{SFCA ($\tau=0.15$)}                             & 76.4 & 75.2 & 74.2 & 73.4 \\
{SFCA ($\tau=0$)}                                & 50.0 & 50.0 & 50.0 & 50.0 \\
\midrule
\multicolumn{5}{l}{\textbf{Ours}} \\
\midrule
{SCA ($\tau=0.3$)}                              & \underline{82.0} & \underline{80.3} & \textbf{79.5} & \textbf{78.4} \\
{SCA ($\tau=0$)}                                & \underline{82.0} & \textbf{80.4} & \textbf{79.5} & \textbf{78.4} \\

\bottomrule
\end{tabular}
\label{tab:multi-answer-auroc-qwen}
\end{table}

\begin{table}[t]
\centering
\small
\setlength{\tabcolsep}{4pt}
\renewcommand{\arraystretch}{0.95}
\caption{
\textbf{AUROC score on {LLaMA-3.1-8B-Instruct}} of different calibration methods under increasing answer mixture.
$\tau=0$ denotes no-filtering, while $\tau\neq0$ uses the threshold value  tuned on the development set.
\textbf{Bold} marks the best, \underline{underlined} the second-best in each column.
$\dagger$ indicates the difference from our SCA method is not statistically significant ($p > 0.05$).
}
\begin{tabular}{lcccc}
\toprule
\textbf{Method} & \textbf{[1]} & \textbf{[1,2]} & \textbf{[1,2,4]} & \textbf{[1,2,4,6]} \\
\midrule
\multicolumn{5}{l}{\textbf{Single-turn}} \\
\midrule
\textit{Question-level} & & & & \\
Prob Entropy                                    & 70.3 & 68.2 & 63.8 & 61.5 \\
N-Prob Entropy                                  & 65.6 & 63.2 & 60.1 & 58.1 \\
Sem Entropy                                     & 78.5 & 77.9 & 74.8 & 73.1 \\
\midrule
\textit{Answer-level} & & & & \\
Verb                                            & 56.6 & 56.9 & 55.1 & 54.4 \\
Verb-Topk                                       & 56.6 & 55.8 & 54.3 & 53.3 \\
Consis                                          & \underline{81.5}$^\dagger$ & 80.7$^\dagger$ & \underline{77.4} & 77.0 \\
Consis-Verb                                     & 77.4 & 76.6 & 74.7 & 74.2 \\
Consis-Verb-Topk                                & 77.8 & 76.6 & 74.8 & 74.1 \\
Perplexity                                      & 80.8$^\dagger$ & 79.3 & 76.2 & 75.4 \\

\midrule
\multicolumn{5}{l}{\textbf{Double-turn}} \\
\midrule
P\textsubscript{True}-Consis                    & 73.5 & 71.1 & 68.1 & 67.6 \\
P\textsubscript{True}-Prob                      & 75.9 & 72.5 & 69.0 & 68.5 \\
P\textsubscript{True}-Consis-Cand               & 73.8 & 71.2 & 68.6 & 67.8 \\
P\textsubscript{True}-Prob-Cand                 & 74.3 & 71.7 & 69.3 & 68.5 \\
Self-Ask                                        & 67.6 & 64.5 & 63.0 & 63.2 \\
Self-Ask-Cand                                   & 77.7 & 75.6 & 73.6 & 72.9 \\
\midrule
\multicolumn{5}{l}{\textbf{Confidence Aggregation Baselines}} \\
\midrule
{SNCA ($\tau=0.2$)}                              & 75.4 & 74.9 & 72.3 & 70.5 \\
{SNCA ($\tau=0$)}                                & 50.0 & 50.0 & 50.0 & 50.0 \\
{SFCA ($\tau=0.1$)}                              & 77.8 & 76.9 & 74.7 & 74.9 \\
{SFCA ($\tau=0$)}                                & 50.0 & 50.0 & 50.0 & 50.0 \\
\midrule
\multicolumn{5}{l}{\textbf{Ours}} \\
\midrule
{SCA ($\tau=0.05$)}                             & \textbf{81.6} & \underline{81.3} & \textbf{78.3} & \underline{77.7} \\
{SCA ($\tau=0$)}                                & \textbf{81.6} & \textbf{81.4} & \textbf{78.3} & \textbf{77.8} \\

\bottomrule
\end{tabular}
\label{tab:multi-answer-auroc-llama8b}
\end{table}

\begin{table}[t]
\caption{\textbf{Confidence scores of} \textbf{Qwen2.5-7B-Instruct} under different ground truth(GT) set sizes.}
\centering
\small
\setlength{\tabcolsep}{3pt}
\renewcommand{\arraystretch}{0.9}
\begin{tabular}{lcccc}
\toprule
\textbf{Method} & \textbf{1a} & \textbf{2a} & \textbf{4a} & \textbf{6a} \\
\midrule
\multicolumn{5}{l}{\textbf{Single-turn}} \\
\midrule
\textit{Question-level} & & & & \\
Prob Entropy & 67.8 & 68.2 & 66.9 & 66.6 \\
N-Prob Entropy & 92.1 & 92.3 & 92.0 & 91.9 \\
Sem Entropy & 64.7 & 65.5 & 63.6 & 62.9 \\
\midrule
\textit{Answer-level} & & & & \\
Verb & 94.7 & 95.4 & 94.8 & 94.7 \\
Verb\textsubscript{Topk} & 82.5 & 81.3 & 73.7 & 71.8 \\
Consistency & 53.1 & 54.7 & 52.6 & 51.9 \\
Consis-Verb & 44.9 & 44.3 & 41.2 & 40.5 \\
Consis-Verb-Topk & 35.1 & 35.3 & 31.1 & 30.7 \\
Perplexity & 79.2 & 79.6 & 79.4 & 79.1 \\
\midrule
\multicolumn{5}{l}{\textbf{Double-turn}} \\
\midrule
P\textsubscript{True}-Consis & 46.9 & 47.0 & 45.0 & 46.9 \\
P\textsubscript{True}-Prob & 46.8 & 46.9 & 44.9 & 46.8 \\
P\textsubscript{True}-Consis-Cand & 59.3 & 61.0 & 62.9 & 65.0 \\
P\textsubscript{True}-Prob-Cand & 59.4 & 61.1 & 63.0 & 65.2 \\
Self\textsubscript{Ask} & 73.7 & 73.8 & 73.0 & 73.7 \\
Self\textsubscript{Ask}-Cand & 70.5 & 71.5 & 71.1 & 72.2 \\
\bottomrule
\end{tabular}
\label{tab:confidence_qwen257b}
\end{table}

\begin{table}[t]
\caption{\textbf{Confidence scores of} \textbf{Qwen2.5-14B-Instruct} under different ground truth(GT) set sizes.}
\centering
\small
\setlength{\tabcolsep}{3pt}
\renewcommand{\arraystretch}{0.9}
\begin{tabular}{lcccc}
\toprule
\textbf{Method} & \textbf{1a} & \textbf{2a} & \textbf{4a} & \textbf{6a} \\
\midrule
\multicolumn{5}{l}{\textbf{Single-turn}} \\
\midrule
\textit{Question-level} & & & & \\
Prob Entropy & 73.4 & 73.3 & 72.4 & 71.7 \\
N-Prob Entropy & 93.7 & 93.6 & 93.5 & 93.2 \\
Sem Entropy & 70.6 & 71.3 & 69.9 & 68.4 \\
\midrule
\textit{Answer-level} & & & & \\
Verb & 87.4 & 87.1 & 85.6 & 85.4 \\
Verb\textsubscript{Topk} & 87.6 & 84.0 & 74.1 & 70.5 \\
Consistency & 62.7 & 63.3 & 62.5 & 61.0 \\
Consis-Verb & 42.2 & 45.6 & 43.4 & 41.7 \\
Consis-Verb-Topk & 45.5 & 44.8 & 42.7 & 40.7 \\
Perplexity & 85.6 & 85.8 & 85.8 & 85.4 \\
\midrule
\multicolumn{5}{l}{\textbf{Double-turn}} \\
\midrule
P\textsubscript{True}-Consis & 50.1 & 52.2 & 49.4 & 49.5 \\
P\textsubscript{True}-Prob & 50.1 & 52.2 & 49.4 & 49.5 \\
P\textsubscript{True}-Consis-Cand & 67.1 & 69.8 & 71.8 & 71.9 \\
P\textsubscript{True}-Prob-Cand & 67.1 & 69.8 & 71.8 & 71.9 \\
Self\textsubscript{Ask} & 77.5 & 78.7 & 76.0 & 75.2 \\
Self\textsubscript{Ask}-Cand & 76.2 & 78.2 & 78.0 & 77.4 \\
\bottomrule
\end{tabular}
\label{tab:confidence_qwen2514b}
\end{table}

\begin{table}[t]
\caption{\textbf{Confidence scores of} \textbf{Qwen2.5-32B-Instruct} under different ground truth(GT) set sizes.}
\centering
\small
\setlength{\tabcolsep}{3pt}
\renewcommand{\arraystretch}{0.9}
\begin{tabular}{lcccc}
\toprule
\textbf{Method} & \textbf{1a} & \textbf{2a} & \textbf{4a} & \textbf{6a} \\
\midrule
\multicolumn{5}{l}{\textbf{Single-turn}} \\
\midrule
\textit{Question-level} & & & & \\
Prob Entropy & 74.9 & 74.2 & 73.4 & 72.7 \\
N-Prob Entropy & 94.1 & 94.1 & 93.8 & 93.6 \\
Sem Entropy & 72.8 & 72.4 & 71.2 & 69.1 \\
\midrule
\textit{Answer-level} & & & & \\
Verb & 88.6 & 88.1 & 87.3 & 87.7 \\
Verb\textsubscript{Topk} & 87.6 & 82.4 & 75.1 & 72.4 \\
Consistency & 64.9 & 63.9 & 62.9 & 61.2 \\
Consis-Verb & 54.1 & 52.7 & 51.8 & 50.5 \\
Consis-Verb-Topk & 48.1 & 45.9 & 44.7 & 42.2 \\
Perplexity & 85.9 & 85.7 & 85.5 & 85.0 \\
\midrule
\multicolumn{5}{l}{\textbf{Double-turn}} \\
\midrule
P\textsubscript{True}-Consis & 69.4 & 76.2 & 78.2 & 78.4 \\
P\textsubscript{True}-Prob & 69.5 & 76.2 & 78.3 & 78.5 \\
P\textsubscript{True}-Consis-Cand & 74.9 & 78.7 & 81.0 & 81.9 \\
P\textsubscript{True}-Prob-Cand & 74.9 & 78.7 & 81.1 & 81.9 \\
Self\textsubscript{Ask} & 72.8 & 72.5 & 70.7 & 72.4 \\
Self\textsubscript{Ask}-Cand & 73.2 & 75.2 & 75.6 & 76.1 \\
\bottomrule
\end{tabular}
\label{tab:confidence_qwen2532b}
\end{table}

\begin{table}[t]
\caption{\textbf{Confidence scores of} \textbf{Qwen2.5-72B-Instruct} under different ground truth(GT) set sizes.}
\centering
\small
\setlength{\tabcolsep}{3pt}
\renewcommand{\arraystretch}{0.9}
\begin{tabular}{lcccc}
\toprule
\textbf{Method} & \textbf{1a} & \textbf{2a} & \textbf{4a} & \textbf{6a} \\
\midrule
\multicolumn{5}{l}{\textbf{Single-turn}} \\
\midrule
\textit{Question-level} & & & & \\
Prob Entropy & 76.4 & 75.4 & 74.3 & 73.4 \\
N-Prob Entropy & 94.1 & 93.9 & 93.5 & 93.3 \\
Sem Entropy & 71.8 & 70.3 & 68.3 & 66.6 \\
\midrule
\textit{Answer-level} & & & & \\
Verb & 87.9 & 87.3 & 86.2 & 86.8 \\
Verb\textsubscript{Topk} & 85.9 & 79.5 & 70.8 & 69.9 \\
Consistency & 65.4 & 64.2 & 61.5 & 59.8 \\
Consis-Verb & 59.0 & 58.2 & 55.5 & 53.7 \\
Consis-Verb-Topk & 53.2 & 52.8 & 48.1 & 45.0 \\
Perplexity & 87.7 & 87.2 & 86.2 & 85.5 \\
\midrule
\multicolumn{5}{l}{\textbf{Double-turn}} \\
\midrule
P\textsubscript{True}-Consis & 75.1 & 73.3 & 68.6 & 73.7 \\
P\textsubscript{True}-Prob & 75.2 & 73.3 & 68.8 & 73.9 \\
P\textsubscript{True}-Consis-Cand & 75.8 & 76.0 & 73.4 & 77.6 \\
P\textsubscript{True}-Prob-Cand & 75.8 & 76.1 & 73.5 & 77.7 \\
Self\textsubscript{Ask} & 84.9 & 79.7 & 72.9 & 75.3 \\
Self\textsubscript{Ask}-Cand & 83.5 & 81.8 & 76.6 & 79.5 \\
\bottomrule
\end{tabular}
\label{tab:confidence_qwen2572b}
\end{table}


\begin{table}[t]
\caption{\textbf{Confidence scores of} \textbf{LLaMA-3.1-8B-Instruct} under different ground truth(GT) set sizes.}
\centering
\small
\setlength{\tabcolsep}{3pt}
\renewcommand{\arraystretch}{0.9}
\begin{tabular}{lcccc}
\toprule
\textbf{Method} & \textbf{1a} & \textbf{2a} & \textbf{4a} & \textbf{6a} \\
\midrule
\multicolumn{5}{l}{\textbf{Single-turn}} \\
\midrule
\textit{Question-level} & & & & \\
Prob Entropy & 63.4 & 62.5 & 60.6 & 59.5 \\
N-Prob Entropy & 89.7 & 89.3 & 88.8 & 88.6 \\
Sem Entropy & 49.9 & 51.4 & 47.5 & 45.1 \\
\midrule
\textit{Answer-level} & & & & \\
Verb & 96.2 & 96.1 & 96.2 & 96.2 \\
Verb\textsubscript{Topk} & 95.8 & 96.3 & 96.2 & 96.3 \\
Consistency & 36.0 & 36.8 & 32.7 & 30.4 \\
Consis-Verb & 37.1 & 37.0 & 33.5 & 30.7 \\
Consis-Verb-Topk & 30.4 & 29.8 & 26.9 & 24.6 \\
Perplexity & 67.0 & 67.6 & 65.9 & 64.4 \\
\midrule
\multicolumn{5}{l}{\textbf{Double-turn}} \\
\midrule
P\textsubscript{True}-Consis & 53.3 & 50.8 & 46.2 & 45.9 \\
P\textsubscript{True}-Prob & 53.1 & 50.9 & 46.4 & 45.7 \\
P\textsubscript{True}-Consis-Cand & 64.8 & 65.7 & 66.1 & 65.7 \\
P\textsubscript{True}-Prob-Cand & 64.7 & 65.7 & 65.9 & 65.9 \\
Self\textsubscript{Ask} & 37.7 & 31.7 & 26.8 & 27.2 \\
Self\textsubscript{Ask}-Cand & 45.4 & 47.8 & 46.3 & 46.1 \\
\bottomrule
\end{tabular}
\label{tab:confidence_llama318b}
\end{table}


\begin{table}[t]
\caption{\textbf{Confidence scores of} \textbf{GPT-4o-mini} under different ground truth(GT) set sizes.}
\centering
\small
\setlength{\tabcolsep}{3pt}
\renewcommand{\arraystretch}{0.9}
\begin{tabular}{lcccc}
\toprule
\textbf{Method} & \textbf{1a} & \textbf{2a} & \textbf{4a} & \textbf{6a} \\
\midrule
\multicolumn{5}{l}{\textbf{Single-turn}} \\
\midrule
\textit{Question-level} & & & & \\
Sem Entropy & 49.6 & 50.1 & 46.2 & 44.9 \\
\midrule
\textit{Answer-level} & & & & \\
Verb & 93.2 & 92.1 & 91.1 & 90.8 \\
Verb\textsubscript{Topk} & 83.7 & 73.0 & 61.5 & 56.5 \\
Consistency & 60.1 & 60.3 & 57.6 & 56.5 \\
Consis-Verb & 54.1 & 54.3 & 51.2 & 50.2 \\
Consis-Verb-Topk & 48.7 & 46.3 & 42.4 & 41.1 \\
\midrule
\multicolumn{5}{l}{\textbf{Double-turn}} \\
\midrule
P\textsubscript{True}-Consis & 62.5 & 67.7 & 68.3 & 70.5 \\
P\textsubscript{True}-Consis-Cand & 72.3 & 77.0 & 77.6 & 79.1 \\
Self\textsubscript{Ask} & 76.3 & 76.1 & 73.7 & 74.9 \\
Self\textsubscript{Ask}-Cand & 73.7 & 76.4 & 76.9 & 76.6 \\
\bottomrule
\end{tabular}
\label{tab:confidence_gpt4omini}
\end{table}

\begin{table}[t]
\caption{\textbf{Confidence scores of} \textbf{GPT-4o} under different ground truth(GT) set sizes.}
\centering
\small
\setlength{\tabcolsep}{3pt}
\renewcommand{\arraystretch}{0.9}
\begin{tabular}{lcccc}
\toprule
\textbf{Method} & \textbf{1a} & \textbf{2a} & \textbf{4a} & \textbf{6a} \\
\midrule
\multicolumn{5}{l}{\textbf{Single-turn}} \\
\midrule
\textit{Question-level} & & & & \\
Sem Entropy & 63.0 & 54.1 & 45.4 & 40.4 \\
\midrule
\textit{Answer-level} & & & & \\
Verb & 92.1 & 90.7 & 88.8 & 88.5 \\
Verb\textsubscript{Topk} & 89.6 & 85.6 & 71.8 & 67.4 \\
Consistency & 70.1 & 64.3 & 56.5 & 52.6 \\
Consis-Verb & 68.8 & 62.8 & 54.8 & 50.6 \\
Consis-Verb-Topk & 65.1 & 58.5 & 48.8 & 45.8 \\
\midrule
\multicolumn{5}{l}{\textbf{Double-turn}} \\
\midrule
P\textsubscript{True}-Consis & 64.3 & 66.8 & 68.4 & 70.4 \\
P\textsubscript{True}-Consis-Cand & 80.0 & 81.0 & 80.5 & 80.0 \\
Self\textsubscript{Ask} & 77.8 & 75.8 & 74.9 & 75.3 \\
Self\textsubscript{Ask}-Cand & 80.5 & 80.4 & 79.0 & 77.9 \\
\bottomrule
\end{tabular}
\label{tab:confidence_gpt4o}
\end{table}


\begin{table}[t]
\caption{\textbf{Confidence scores of} \textbf{DeepSeek-V3} under different ground truth(GT) set sizes.}
\centering
\small
\setlength{\tabcolsep}{3pt}
\renewcommand{\arraystretch}{0.9}
\begin{tabular}{lcccc}
\toprule
\textbf{Method} & \textbf{1a} & \textbf{2a} & \textbf{4a} & \textbf{6a} \\
\midrule
\multicolumn{5}{l}{\textbf{Single-turn}} \\
\midrule
\textit{Question-level} & & & & \\
Sem Entropy & 80.9 & 78.6 & 76.2 & 72.9 \\
\midrule
\textit{Answer-level} & & & & \\
Verb & 85.3 & 85.5 & 85.8 & 86.0 \\
Verb\textsubscript{Topk} & 63.5 & 57.5 & 52.7 & 50.5 \\
Consistency & 81.1 & 79.1 & 76.7 & 73.2 \\
Consis-Verb & 69.0 & 65.9 & 60.6 & 55.6 \\
Consis-Verb-Topk & 63.4 & 58.5 & 53.6 & 49.0 \\
\midrule
\multicolumn{5}{l}{\textbf{Double-turn}} \\
\midrule
P\textsubscript{True}-Consis & 64.3 & 64.6 & 64.7 & 66.2 \\
P\textsubscript{True}-Consis-Cand & 82.4 & 81.5 & 82.5 & 80.5 \\
Self\textsubscript{Ask} & 76.5 & 72.8 & 69.8 & 71.2 \\
Self\textsubscript{Ask}-Cand & 83.8 & 81.3 & 80.8 & 78.7 \\
\bottomrule
\end{tabular}
\label{tab:confidence_deepseekv3}
\end{table}

\end{document}